\documentclass[10pt,twocolumn,letterpaper]{article}

\usepackage{iccv}
\usepackage{times}
\usepackage{epsfig}
\usepackage{graphicx}
\usepackage{amsmath}
\usepackage{amssymb}

\usepackage{xcolor}
\usepackage{caption}
\usepackage{subcaption}
\usepackage{tabularx}
\usepackage{rotating}
\usepackage{verbatim}
\usepackage{xcolor,colortbl}
\usepackage{etoolbox}
\usepackage[normalem]{ulem}
\usepackage{booktabs}
\usepackage{multirow}
\usepackage{lipsum}
\usepackage{stmaryrd}
\usepackage{stackengine}
\usepackage{makecell}
\usepackage{pifont}
\usepackage{cancel}
\usepackage{adjustbox}
\usepackage{dblfloatfix}

\usepackage[pagebackref=true,breaklinks=true,letterpaper=true,colorlinks,bookmarks=false]{hyperref}
\hypersetup{colorlinks}

\newcommand{\cmark}{\ding{51}}%

\usepackage[capitalize]{cleveref}
\crefname{section}{Sec.}{Secs.}
\Crefname{section}{Section}{Sections}
\Crefname{table}{Table}{Tables}
\crefname{table}{Tab.}{Tabs.}

\definecolor{car}{rgb}{0.39215686, 0.58823529, 0.96078431}
\definecolor{bicycle}{rgb}{0.39215686, 0.90196078, 0.96078431}
\definecolor{motorcycle}{rgb}{0.11764706, 0.23529412, 0.58823529}
\definecolor{truck}{rgb}{0.31372549, 0.11764706, 0.70588235}
\definecolor{other-vehicle}{rgb}{0.39215686, 0.31372549, 0.98039216}
\definecolor{person}{rgb}{1.        , 0.11764706, 0.11764706}
\definecolor{bicyclist}{rgb}{1.        , 0.15686275, 0.78431373}
\definecolor{motorcyclist}{rgb}{0.58823529, 0.11764706, 0.35294118}
\definecolor{road}{rgb}{1.        , 0.        , 1.        }
\definecolor{parking}{rgb}{1.        , 0.58823529, 1.        }
\definecolor{sidewalk}{rgb}{0.29411765, 0.        , 0.29411765}
\definecolor{other-ground}{rgb}{0.68627451, 0.        , 0.29411765}
\definecolor{building}{rgb}{1.        , 0.78431373, 0.        }
\definecolor{fence}{rgb}{1.        , 0.47058824, 0.19607843}
\definecolor{vegetation}{rgb}{0.        , 0.68627451, 0.        }
\definecolor{trunk}{rgb}{0.52941176, 0.23529412, 0.        }
\definecolor{terrain}{rgb}{0.58823529, 0.94117647, 0.31372549}
\definecolor{pole}{rgb}{1.        , 0.94117647, 0.58823529}
\definecolor{traffic-sign}{rgb}{1.        , 0.        , 0.    }

\makeatletter
\newcommand{\car@semkitfreq}{3.92}
\newcommand{\bicycle@semkitfreq}{0.03}
\newcommand{\motorcycle@semkitfreq}{0.03}
\newcommand{\truck@semkitfreq}{0.16}
\newcommand{\othervehicle@semkitfreq}{0.20}
\newcommand{\person@semkitfreq}{0.07}
\newcommand{\bicyclist@semkitfreq}{0.07}
\newcommand{\motorcyclist@semkitfreq}{0.05}
\newcommand{\road@semkitfreq}{15.30}  %
\newcommand{\parking@semkitfreq}{1.12}
\newcommand{\sidewalk@semkitfreq}{11.13}  %
\newcommand{\otherground@semkitfreq}{0.56}
\newcommand{\building@semkitfreq}{14.1}  %
\newcommand{\fence@semkitfreq}{3.90}
\newcommand{\vegetation@semkitfreq}{39.3}  %
\newcommand{\trunk@semkitfreq}{0.51}
\newcommand{\terrain@semkitfreq}{9.17} %
\newcommand{\pole@semkitfreq}{0.29}
\newcommand{\trafficsign@semkitfreq}{0.08}
\newcommand{\semkitfreq}[1]{{\csname #1@semkitfreq\endcsname}}

\iccvfinalcopy

\ificcvfinal\fi

\begin{document}

\title{OccFormer: Dual-path Transformer for Vision-based\\3D Semantic Occupancy Prediction}
\author{
Yunpeng Zhang\\
PhiGent Robotics\\
{\tt\small yunpengzhang97@gmail.com}
\and
Zheng Zhu\thanks{Corresponding author.}\\
PhiGent Robotics\\
{\tt\small zhengzhu@ieee.org}
\and
Dalong Du\\
PhiGent Robotics\\
{\tt\small dalong.du@phigent.ai}
}

\maketitle
\thispagestyle{empty}

\begin{abstract}
The vision-based perception for autonomous driving has undergone a transformation from the bird-eye-view (BEV) representations to the 3D semantic occupancy. Compared with the BEV planes, the 3D semantic occupancy further provides structural information along the vertical direction. This paper presents OccFormer, a dual-path transformer network to effectively process the 3D volume for semantic occupancy prediction. OccFormer achieves a long-range, dynamic, and efficient encoding of the camera-generated 3D voxel features. It is obtained by decomposing the heavy 3D processing into the local and global transformer pathways along the horizontal plane. For the occupancy decoder, we adapt the vanilla Mask2Former for 3D semantic occupancy by proposing preserve-pooling and class-guided sampling, which notably mitigate the sparsity and class imbalance. Experimental results demonstrate that OccFormer significantly outperforms existing methods for semantic scene completion on SemanticKITTI dataset and for LiDAR semantic segmentation on nuScenes dataset. Code is available at \url{https://github.com/zhangyp15/OccFormer}.
\end{abstract}
\vspace{-3mm}
\section{Introduction}
The accurate perception of 3D surroundings constitutes the foundation of modern autonomous driving systems. Though LiDAR-based methods~\cite{pointpillars, pointrcnn, cylindrical++, roldao2020lmscnet, F-pointnet, centertrack}, with explicit depth measurements, have been dominating the leading performance on public datasets~\cite{kitti, nuscenes, waymo, behley2019semantickitti}, vision-based approaches still offer advantages in terms of cost-effectiveness, stability, and generality. 
The past years have witnessed the prosperity of Bird-Eye-View representations for vision-based 3D perception. With the multi-view camera images as input, various attempts for 2D-to-3D transformation~\cite{lss, BEVFormer, bevdet, bevdepth} have been proposed for applications including 3D object detection~\cite{bevdet, BEVFormer, PETR}, semantic map construction~\cite{lss, CVT, pon, bevsegformer}, and motion prediction~\cite{hu2021fiery, stretchbev, zhang2022beverse}. 
Considering these tasks require either rigid bounding boxes or BEV-oriented predictions, the collapse of 3D scenes into 2D ground planes has demonstrated an excellent trade-off between performance and efficiency. However, the holistic understanding of the 3D scene, especially for real-world obstacles with variable shapes, can hardly be recovered with the condensed BEV feature maps. To this end, this paper focuses on building a fine-grained 3D representation, namely 3D semantic occupancy, for the surrounding environment with multi-view images. 

The task of 3D semantic occupancy prediction aims to reconstruct the surrounding 3D environment with fine-grained geometry and semantics, which is also known as 3D semantic scene completion when the LiDAR point cloud is taken as input. For the driving scenes, most existing methods~\cite{roldao2020lmscnet, cheng2021s3cnet, 3d-sketch, li2020anisotropic, JS3CNet} still rely on the expensive LiDAR sensors for explicit depth measurements.
The seminar work MonoScene~\cite{cao2022monoscene} proposed the first monocular framework for 3D semantic occupancy prediction. It first constructs the 3D feature with sight projection and then processes it with a classical 3D UNet. However, the 3D convolution suffers from several limitations. First, it reasons the semantics within a relatively fixed receptive field, while different semantic classes may distribute following various patterns. Also, its spatial invariance cannot well process the sparse and discontinuous 3D features, generated from the state-of-the-art practices for image-to-3D transformation~\cite{lss, bevdet, bevdepth}. 
Finally, the 3D convolution filters can consume massive parameters. Therefore, we believe a long-range, dynamic, and efficient method for encoding 3D features is needed to pave the way.

Inspired by the widespread success of vision transformers~\cite{vit, swin_transformer} in various vision tasks~\cite{detr, setr, action_transformer, pointr, swin_transformer, vitdet}, we are motivated to utilize the attention mechanism for building the encoder-decoder network for 3D semantic occupancy prediction. 
For the encoder part, we propose the dual-path transformer block to unleash the capacity of self-attention while limiting the quadratic complexity. Specifically, the local path operates along each 2D BEV slice with the shared windowed attention to capture the fine-grained details, while the global path performs on the collapsed BEV feature to obtain scene-level understanding. Finally, the dual-path outputs are adaptively fused to generate the output 3D feature volume. The dual-path designs appropriately break down the challenging processing of 3D feature volumes and we demonstrate its clear advantage over the classic 3D convolutions. 
For the decoder part, we are the first to adapt the state-of-the-art method Mask2Former~\cite{mask2former} for 3D semantic occupancy prediction. We further propose to use max-pooling rather than the default bilinear for computing the masked regions for attention, which can better preserve the minor classes. Additionally, the class-guided sampling is proposed to capture the foreground areas for more effective optimization.
Experimental results demonstrate the superiority of OccFormer over existing state-of-the-art methods. For 3D semantic scene completion on SemanticKITTI~\cite{behley2019semantickitti} dataset, OccFormer outperforms MonoScene by 1.24\% mIoU, which makes an 11\% relative improvement and ranks first on the test leaderboard among all monocular methods. We also evaluate OccFormer on nuScenes~\cite{nuscenes} dataset for LiDAR semantic segmentation, following TPVFormer~\cite{tpvformer}. Our method surpasses TPVFormer by 1.4\% mIoU and generates more complete and realistic predictions for 3D semantic occupancy prediction. 
\section{Related Work}

\subsection{Camera-based BEV Perception}
Considering the dimension gap between the 2D image input and the 3D prediction, recent studies for vision-based 3D perception first construct the BEV feature representations and then perform various downstream tasks on the BEV space~\cite{bevdet, bevdepth, BEVFormer, bevsegformer, zhang2022beverse, lss, CVT, caddn, hu2021fiery, stretchbev, pon}. 
To transform the perspective image features into the BEV features, LSS~\cite{lss} and its follow-ups~\cite{caddn, bevdepth, hu2021fiery, zhang2022beverse} predict the pixel-wise depth distribution to project the image features into 3D points, which are then voxelized into the BEV features. Other methods like BEVFormer~\cite{BEVFormer} utilize the deformable attention~\cite{deformable_DETR, transformer} to update the BEV queries with corresponding image features. 
In this paper, we extend the BEV-based perception to 3D semantic occupancy prediction, which further contains the structural information along the height dimension.

\subsection{3D Semantic Occupancy Prediction}
Since 3D semantic occupancy prediction is also known as 3D semantic scene completion (SSC), we also review the related SSC methods. 
SSCNet~\cite{sscnet} first proposes the problem of semantic scene completion, which jointly reasons the geometry and semantics. The follow-ups usually employ the geometrical inputs with explicit depth information~\cite{roldao2020lmscnet, li2020anisotropic, JS3CNet, cheng2021s3cnet, 3d-sketch, local_diffs}. 
Recently, MonoScene~\cite{cao2022monoscene} builds the first monocular method for semantic scene completion, which employs the 3D UNet to process the voxel features generated by sight projection. TPVFormer~\cite{tpvformer} proposes the tri-perspective view representation to describe the 3D scene for semantic occupancy prediction. 
Despite its simplicity, the tri-plane format is susceptible to the deficiency of fine-grained semantic information, leading to inferior performance. In this paper, we re-advocate the representation power of dense 3D features and propose the transformer-based encoder-decoder network for 3D semantic occupancy prediction. 

\vspace{-1mm}
\subsection{Efficient 3D Network}
On the field of 3D semantic scene completion, extensive attempts have been proposed to improve the efficiency of 3D networks. EsscNet~\cite{EsscNet} partitions the non-empty voxels into different groups and conduct 3D sparse convolution within each group. 
DDRNet~\cite{DDR-SSC} replaces the 3D convolution with three consecutive 1D convolution layers along each dimension. 
AIC-Net~\cite{li2020anisotropic} further equips each 1D layer with various kernel sizes for anisotropic processing. LMSCNet~\cite{roldao2020lmscnet} uses the 2D UNet to process the collapsed BEV features and finally expands the height dimension for 3D segmentation. 
S3CNet~\cite{cheng2021s3cnet} turns to the sparse convolution for outdoor point clouds. 
These methods are mostly targeted for LiDAR points with convolutional structures. In this paper, we propose the dual-path transformer to efficiently process the camera-generated 3D feature volumes with transformer-based modules. 

\section{Approach}

\begin{figure*}[t]
\centering
\includegraphics[width=\linewidth]{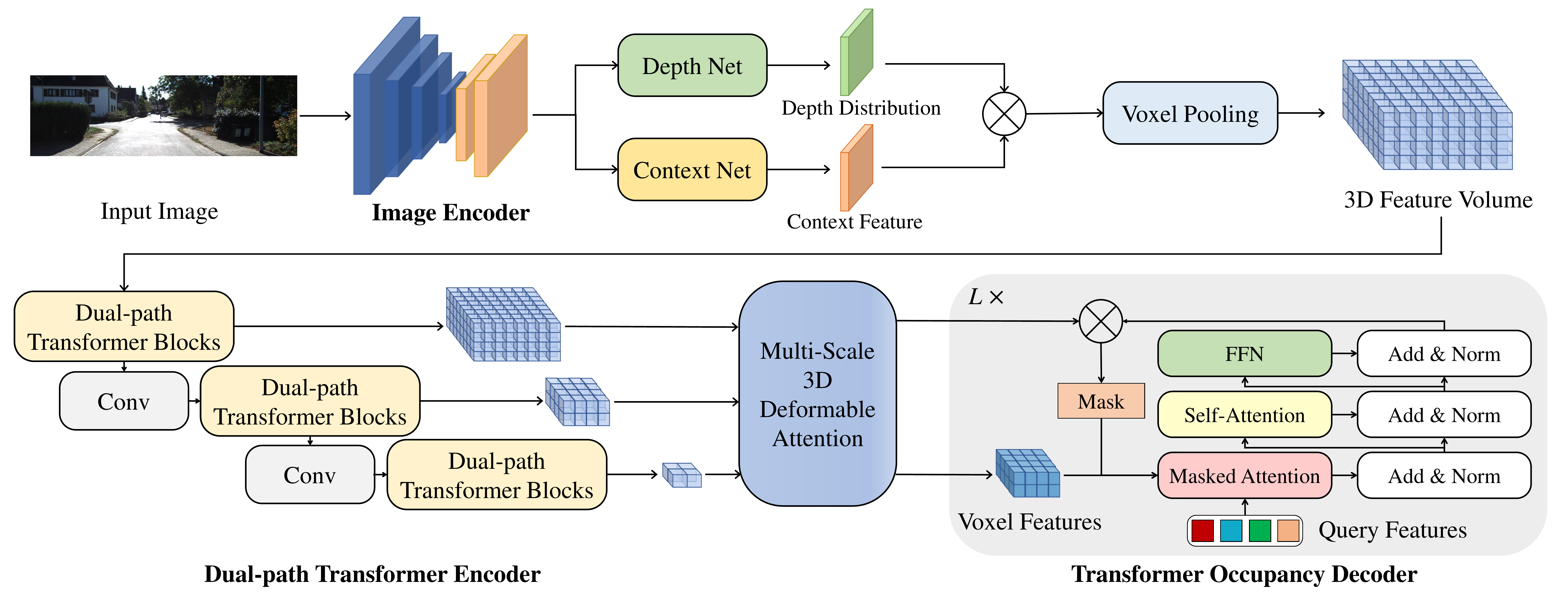}
\vspace{-6mm}
\caption{The framework of the proposed OccFormer for camera-based 3D semantic occupancy prediction. The pipeline consists of the image encoder for extracting multi-scale 2D features, the image-to-3D transformation for lifting the 2D features to 3D volumes, and the transformer-based encoder-decoder for obtaining 3D semantic features and predicting the 3D semantic occupancy.}
\label{fig:framework}
\vspace{-4mm}
\end{figure*}

\subsection{Overview}
The overall pipeline of OccFormer is illustrated in~\cref{fig:framework}. With the monocular image or multi-camera images as the input, the multi-scale features are first extracted by the image encoder, and then lifted to 3D feature volume, which are briefly introduced in the following paragraphs. 
The 3D feature is further processed by the dual-path transformer encoder (\cref{sec:dual_path_transformer_encoder}) to produce multi-scale voxel features with local and global semantics. Finally, the transformer occupancy decoder (\cref{sec:transformer_occupancy_decoder}) fuses multi-scale features and formulates the occupancy prediction as the transformer-based mask classification for decoding.

\vspace{-3mm}
\paragraph{Image Encoder.}
The image encoder aims to extract geometric and semantic features within the perspective view, which provides the foundation of the later-generated 3D feature volume. The image encoder consists of a backbone network for extracting multi-scale features and a neck for further fusion. 
The output of the image encoder is one fused feature map with $\frac{1}{16}$ of the input resolution. We use $\mathbf{F}^{2d} \in \mathbb{R}^{N \times C \times H \times W}$ to represent the extracted features, where $N$ is the number of camera views, $C$ is the channel number, and $\left(H, W\right)$ refers to the resolution. 

\vspace{-3mm}
\paragraph{Image-to-3D Transformation.}
Inspired by recent studies on lifting multi-view images to the Bird-Eye-View representations~\cite{lss,bevdet,bevdepth,BEVFormer}, we extend the LSS~\cite{lss} paradigm for image-to-3D transformation. Specifically, the encoded image features $\mathbf{F}^{2d}$ are processed to generate the context feature $\mathbf{F}^{2d}_{con} \in \mathbb{R}^{N \times C_{con} \times H \times W}$ and the discrete depth distribution $\mathbf{D} \in \mathbb{R}^{N \times D \times H \times W}$. 
Then the outer product $\mathbf{F}^{2d}_{con} \otimes \mathbf{D}$ is employed to create the point cloud representation $\mathbf{P} \in \mathbb{R}^{NDHW\times C_{con}}$. Finally, the voxel-pooling is conducted to create the 3D feature volume $\mathbf{F}^{3d} \in \mathbb{R}^{C_{con} \cdot X \cdot Y \cdot Z}$, where $\left(X, Y, Z\right)$ denotes the resolution of the 3D volume. 

\subsection{Dual-path Transformer Encoder}
\label{sec:dual_path_transformer_encoder}

\begin{figure}[t]
\centering
\includegraphics[width=\linewidth]{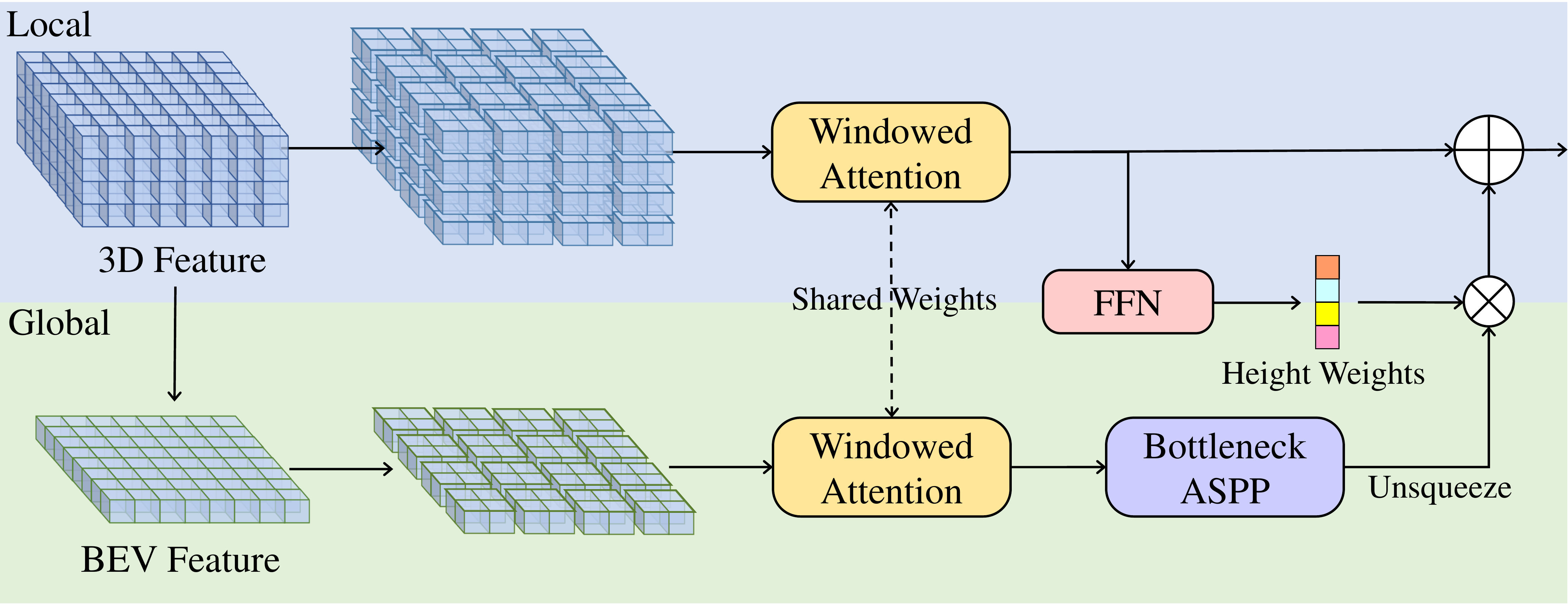}
\vspace{-4mm}
\caption{Illustration of the dual-path transformer block. The local path processes the 3D feature by applying the shared windowed attention to each horizontal slice, while the global path operates on the collapsed BEV feature for scene-level semantic layouts. The dual-path outputs are finally fused through the weighted summation. The skip connection is omitted.}
\vspace{-4mm}
\label{fig:dualpath_block}
\end{figure}

To pursue long-range, dynamic, and efficient processing of the 3D feature volumes, we propose the dual-path transformer block to build the 3D encoder. 
Inspired by recent advances that introduce locality into the transformer~\cite{vitae, vitaev2, mpvit}, we also design the encoder as a hybrid structure. The encoder consists of a series of dual-path transformer blocks, while one 3D convolution layer is inserted between two consecutive blocks to introduce locality and optionally perform the downsampling. The detailed structure of the dual-path transformer block is shown in~\cref{fig:dualpath_block}. With the input 3D feature, the local and global pathways first aggregate semantic information along the horizontal direction in parallel. Next, the dual-path outputs are fused through the sigmoid-weighted summation. Finally, the skip connection is applied to ensure the residual learning~\cite{resnet}. We introduce the dual-path processing with more details in the following paragraph.

The local path is mainly targeted to extract the fine-grained semantic structures. Since the horizontal direction contains the most variations, we believe the parallel processing of all BEV slices with one shared encoder is able to keep most of the semantic information. Specifically, we merge the height dimension into the batch dimension and employ the windowed self-attention~\cite{swin_transformer} as the local feature extractor, which can dynamically attend to long-range regions with moderate computations. 
On the other hand, the global path aims to efficiently capture the scene-level semantic layouts. To this end, the global path starts by getting the BEV feature by average pooling along the height dimension. The same windowed self-attention from the local path is utilized to process the BEV feature for neighbouring semantics. Since we find the global self-attention on the BEV plane can consume excessive memories, the ASPP~\cite{deeplab} is applied instead to capture the global contexts. In practice, we employ the bottleneck structure~\cite{resnet} to reduce the channel number by 4$\times$ for ASPP.
Finally, the scene-level information from the global path is propagated to the entire 3D volume from the local path. Assume the dual-path outputs are $\mathbf{F}_\text{local} \in \mathbb{R}^{C \cdot X \cdot Y \cdot Z}$ and $\mathbf{F}_\text{global} \in \mathbb{R}^{C \cdot X \cdot Y}$, the combined output $\mathbf{F}_\text{out}$ is computed as:
\begin{equation}
    \mathbf{F}_\text{out} = \mathbf{F}_\text{local} + \sigma(\mathbf{W} \mathbf{F}_\text{local}) \cdot \text{unsqueeze} (\mathbf{F}_\text{global}, -1)
\end{equation}
where $\mathbf{W}$ refers to the FFN for generating the aggregation weights along the height dimension, $\sigma(\cdot)$ is the sigmoid function, and ``unsqueeze'' expands the global 2D feature along the height. 
Although the dual-path processing only performs 2D reasoning along the horizontal direction, their combination effectively aggregates essential information for semantic reasoning, including local semantic structures and global semantic layouts. Additionally, the dual-path transformer encoder has fewer parameters and requires less computation than classic 3D convolutions, benefiting from shared modules and mostly 2D reasoning.

\subsection{Transformer Occupancy Decoder}
\label{sec:transformer_occupancy_decoder}
Inspired by the recent mask classification models~\cite{maskformer, mask2former} for image segmentation, we also formulate the 3D semantic occupancy as predicting a set of binary 3D masks associated with corresponding class labels. 
Following Mask2Former~\cite{mask2former}, our transformer occupancy decoder includes the pixel decoder (\cref{sec:pixel_decoder}) for per-voxel embeddings and the transformer decoder (\cref{sec:transformer_decoder}) for per-query embeddings and class predictions. The final mask predictions are derived from the dot product between these two embeddings. 
Also, we introduce two essential modifications to effectively improve the occupancy predictions, including the preserve-pooling (\cref{sec:preserve_pooling}) and the class-guided sampling (\cref{sec:class_guided_sampling}). 
Formally, the input multi-scale feature volumes from the transformer encoder are denoted as $\{ \mathbf{F}^{3d}_i \in \mathbb{R}^{C_i \cdot X_i \cdot Y_i \cdot Z_i} \}_{i=1}^{N_{l}} $, where $N_{l}$ is the level number, $C_i$ is the channel number, and $(X_i, Y_i, Z_i)$ is the volume size. 

\vspace{-3mm}
\subsubsection{Pixel Decoder}
\label{sec:pixel_decoder}

With multi-scale 3D features as input, the pixel decoder is tasked with aggregating multi-level semantics and creating high-resolution voxel embeddings. Since each feature level places different emphasis on low-level details and high-level semantics, we employ the multi-scale deformable attention~\cite{deformable_DETR}, tailored for 3D, to facilitate effective intra-scale and inter-scale interactions. Take the level-$i$ feature $\mathbf{F}^{3d}_i$ as an example, its corresponding real-world coordinates $\mathbf{P}^{3d}_i \in \mathbb{R}^{X_i \cdot Y_i \cdot Z_i \cdot 3}$ are first computed. Then the features are processed to create the sampling offsets $\mathbf{\Delta}^{3d}_j$ and the attention weights $\mathbf{W}^{3d}_j$ for all levels $j=1,\cdots,N_l$. Finally, the updating process is formulated as in~\cref{equ:MulDeformAttn3D}:
\vspace{-1mm}
\begin{equation}
    \mathbf{F}^{3d}_i = \mathbf{F}^{3d}_i + \sum_{j=1}^{N_l} \left[ \mathbf{W}^{3d}_j \mathbf{F}^{3d}_j \left(\mathbf{P}^{3d}_i + \mathbf{\Delta}^{3d}_j \right) \right]
    \label{equ:MulDeformAttn3D}
\end{equation}
where $\mathbf{F}^{3d} \left(\mathbf{P}^{3d} + \mathbf{\Delta}^{3d}\right)$ conducts the trilinear feature sampling at the corresponding positions.
With the above interactions, each processed feature volume is enhanced by the multi-scale semantic information, which facilitates the following transformer decoder. The feature volume with the highest resolution is projected to generate the per-voxel embeddings $\mathcal{E}_\text{voxel} \in \mathbb{R}^{C_{\mathcal{E}} \cdot X \cdot Y \cdot Z}$, where $C_{\mathcal{E}}$ is the embedding dimension.

\vspace{-3mm}
\subsubsection{Transformer Decoder}
\label{sec:transformer_decoder}
With the input multi-scale voxel features and the parameterized query features, the transformer decoder performs an iterative updating of the query features towards the desired class segments, as shown in~\cref{fig:framework}. Within each iteration layer $l$, the queries features $\mathbf{Q}_l$ first attends to their corresponding foreground regions through the masked attention: 
\begin{equation}
\footnotesize
\mathbf{Q}_{l + 1} = \text{softmax} \left[\mathcal{M}_{l-1} + \mathbf{W}_q \mathbf{Q}_l \left(\mathbf{W}_k\mathbf{F}^{3d}_l \right)^T \right] \mathbf{W}_v \mathbf{F}^{3d}_l + \mathbf{Q}_l
\end{equation}
where $\mathbf{F}^{3d}_l$ is the 3D voxel feature, $\mathcal{M}_{l-1}$ is the attention mask from the previous layer, and $\left(\mathbf{W}_q, \mathbf{W}_k, \mathbf{W}_v\right)$ are linear projection layers. The self-attention is then conducted to exchange context information, followed by the FFN for feature projection. 
At the end of each iteration, each query feature $\mathbf{q}_i$ is projected to predict its semantic logits $\mathbf{p}_i$ and the mask embedding $\mathcal{E}_{\text{mask}_i}$. The latter is further transformed into the binary 3D mask $\mathbf{M}_i$ by a dot product with the per-voxel embeddings $\mathcal{E}_\text{voxel}$ and a sigmoid function. The final 3D semantic occupancy prediction $\mathbf{Y}$ is formulated as:
\vspace{-2mm}
\begin{equation}
    \mathbf{Y} = \sum_{i=1}^{N_q} \mathbf{p}_i \cdot \mathbf{M}_i
    \label{equ:mask2former_inference}
\end{equation}
\vspace{-1mm}
where $N_q$ is the number of query features. 

\begin{table*}
    \footnotesize
    \setlength{\tabcolsep}{0.003\linewidth}
    \caption{\textbf{Semantic scene completion results on SemanticKITTI test set.} * represents these methods are adapted for the RGB inputs, which are implemented and reported in MonoScene~\cite{cao2022monoscene}. Our method outperforms all existing monocular methods for semantic scene completion in both the SC IoU and the SSC mIoU.}
    \vspace{-1mm}
    \newcommand{\classfreq}[1]{{~\tiny(\semkitfreq{#1}\%)}}  %
    \centering
    \begin{tabular}{l|c|c c | c c c c c c c c c c c c c c c c c c c}
    \toprule
    Method
    & \makecell{Input\\ Modality}
    & \makecell{SC\\ IoU} & \makecell{SSC \\ mIoU}
    & \rotatebox{90}{road}
    \rotatebox{90}{\ \ \ \classfreq{road}} 
    & \rotatebox{90}{sidewalk}
    \rotatebox{90}{\ \ \ \classfreq{sidewalk}}
    & \rotatebox{90}{parking}
    \rotatebox{90}{\ \ \ \classfreq{parking}} 
    & \rotatebox{90}{other-grnd}
    \rotatebox{90}{\ \ \ \classfreq{otherground}} 
    & \rotatebox{90}{ building}
    \rotatebox{90}{\ \ \ \classfreq{building}} 
    & \rotatebox{90}{ car}
    \rotatebox{90}{\ \ \ \classfreq{car}} 
    & \rotatebox{90}{ truck}
    \rotatebox{90}{\ \ \ \classfreq{truck}} 
    & \rotatebox{90}{ bicycle}
    \rotatebox{90}{\ \ \ \classfreq{bicycle}} 
    & \rotatebox{90}{motorcycle}
    \rotatebox{90}{\ \ \ \classfreq{motorcycle}} 
    & \rotatebox{90}{ other-veh.}
    \rotatebox{90}{\ \ \  \classfreq{othervehicle}} 
    & \rotatebox{90}{vegetation}
    \rotatebox{90}{\ \ \ \classfreq{vegetation}} 
    & \rotatebox{90}{ trunk}
    \rotatebox{90}{\ \ \ \classfreq{trunk}} 
    & \rotatebox{90}{terrain}
    \rotatebox{90}{\ \ \ \classfreq{terrain}} 
    & \rotatebox{90}{ person}
    \rotatebox{90}{\ \ \ \classfreq{person}} 
    & \rotatebox{90}{ bicyclist}
    \rotatebox{90}{\ \ \ \classfreq{bicyclist}} 
    & \rotatebox{90}{ motorcyclist.}
    \rotatebox{90}{\ \ \ \classfreq{motorcyclist}} 
    & \rotatebox{90}{ fence}
    \rotatebox{90}{\ \ \ \classfreq{fence}} 
    & \rotatebox{90}{ pole}
    \rotatebox{90}{\ \ \ \classfreq{pole}} 
    & \rotatebox{90}{traf.-sign}
    \rotatebox{90}{\ \ \ \classfreq{trafficsign}} 
    \\
    \midrule
    LMSCNet*~\cite{roldao2020lmscnet} & Camera &  31.38 & 7.07 &  46.70 & 19.50 & 13.50 & 3.10 & 10.30 & 14.30 & 0.30 & 0.00 & 0.00 & 0.00 & 10.80 & 0.00 & 10.40 & 0.00 & 0.00 & 0.00 & 5.40 & 0.00 & 0.00   \\
    
    3DSketch*~\cite{3d-sketch} & Camera & 26.85 & 6.23 & 37.70 & 19.80 & 0.00 & 0.00 & 12.10 & 17.10 & 0.00 & 0.00 & 0.00 & 0.00 & 12.10 & 0.00 & 16.10 & 0.00 & 0.00 & 0.00 & 3.40 & 0.00 & 0.00  \\
    
    AICNet*~\cite{li2020anisotropic} & Camera & 23.93 & 7.09	& 39.30	& 18.30 & 19.80 & 1.60 & 9.60	& 15.30	& 0.70	& 0.00	& 0.00	& 0.00	& 9.60	& 1.90	& 13.50	& 0.00	& 0.00	& 0.00	& 5.00	& 0.10	& 0.00 \\
    
    JS3C-Net*~\cite{JS3CNet} & Camera & 34.00 & 8.97 & 47.30 & 21.70 & 19.90 & 2.80 & 12.70 & 20.10 & 0.80 & 0.00 & 0.00 & 4.10 & 14.20 & 3.10 & 12.40 & 0.00 & 0.20 & 0.20 & 8.70 & 1.90 & 0.30  \\
    
    MonoScene~\cite{cao2022monoscene} & Camera & 34.16 & 11.08 & 54.70 & 27.10 & 24.80 & 5.70 & 14.40 & 18.80 & 3.30 & 0.50 & 0.70 & \textbf{4.40} & 14.90 & 2.40 & 19.50 & 1.00 & 1.40 & \textbf{0.40} & 11.10 & 3.30 & 2.10  \\
    
    TPVFormer~\cite{tpvformer} & Camera & 34.25 & 11.26 & 55.10 & 27.20 & 27.40 & \textbf{6.50} & 14.80 & 19.20 & \textbf{3.70} & 1.00 & 0.50 & 2.30 & 13.90 & 2.60 & 20.40 & 1.10 & \textbf{2.40} & 0.30 & 11.00 & 2.90 & 1.50  \\

    OccFormer (ours) & Camera & \textbf{34.53} & \textbf{12.32} & \textbf{55.90} & \textbf{30.30} & \textbf{31.50} & \textbf{6.50} & \textbf{15.70} & \textbf{21.60} & 1.20 & \textbf{1.50} & \textbf{1.70} & 3.20 & \textbf{16.80} & \textbf{3.90} & \textbf{21.30} & \textbf{2.20} & 1.10 & 0.20 & \textbf{11.90} & \textbf{3.80} & \textbf{3.70} \\
    \bottomrule
    \end{tabular}
    \label{table:kitti_test_perf}
    \vspace{-2mm}
\end{table*}

\begin{table*}
    \scriptsize
    \setlength{\tabcolsep}{0.0035\linewidth}
    \caption{\textbf{Semantic scene completion results on SemanticKITTI~\cite{behley2019semantickitti} validation set.} * represents these methods are adapted for the RGB inputs, which are implemented and reported in MonoScene~\cite{cao2022monoscene}. $\dagger$ represents the reproduced result from \cite{tpvformer}.}
    \vspace{-2mm}
    \newcommand{\classfreq}[1]{{~\tiny(\semkitfreq{#1}\%)}}  %
    \centering
    \begin{tabular}{l|c|c|c c c c c c c c c c c c c c c c c c c|c}
        \toprule
        & & SC & \multicolumn{20}{c}{SSC} \\
        Method & SSC Input & IoU
        & \rotatebox{90}{\textcolor{road}{$\blacksquare$} road\classfreq{road}} 
        & \rotatebox{90}{\textcolor{sidewalk}{$\blacksquare$} sidewalk\classfreq{sidewalk}}
        & \rotatebox{90}{\textcolor{parking}{$\blacksquare$} parking\classfreq{parking}} 
        & \rotatebox{90}{\textcolor{other-ground}{$\blacksquare$} other-ground\classfreq{otherground}} 
        & \rotatebox{90}{\textcolor{building}{$\blacksquare$} building\classfreq{building}} 
        & \rotatebox{90}{\textcolor{car}{$\blacksquare$} car\classfreq{car}} 
        & \rotatebox{90}{\textcolor{truck}{$\blacksquare$} truck\classfreq{truck}} 
        & \rotatebox{90}{\textcolor{bicycle}{$\blacksquare$} bicycle\classfreq{bicycle}} 
        & \rotatebox{90}{\textcolor{motorcycle}{$\blacksquare$} motorcycle\classfreq{motorcycle}} 
        & \rotatebox{90}{\textcolor{other-vehicle}{$\blacksquare$} other-vehicle\classfreq{othervehicle}} 
        & \rotatebox{90}{\textcolor{vegetation}{$\blacksquare$} vegetation\classfreq{vegetation}} 
        & \rotatebox{90}{\textcolor{trunk}{$\blacksquare$} trunk\classfreq{trunk}} 
        & \rotatebox{90}{\textcolor{terrain}{$\blacksquare$} terrain\classfreq{terrain}} 
        & \rotatebox{90}{\textcolor{person}{$\blacksquare$} person\classfreq{person}} 
        & \rotatebox{90}{\textcolor{bicyclist}{$\blacksquare$} bicyclist\classfreq{bicyclist}} 
        & \rotatebox{90}{\textcolor{motorcyclist}{$\blacksquare$} motorcyclist\classfreq{motorcyclist}} 
        & \rotatebox{90}{\textcolor{fence}{$\blacksquare$} fence\classfreq{fence}} 
        & \rotatebox{90}{\textcolor{pole}{$\blacksquare$} pole\classfreq{pole}} 
        & \rotatebox{90}{\textcolor{traffic-sign}{$\blacksquare$} traffic-sign\classfreq{trafficsign}} 
        & mIoU\\
        \midrule
        LMSCNet*~\cite{roldao2020lmscnet} & $\hat{x}^{\text{occ}}_{\text{3D}}$ & 28.61 & 40.68 & 18.22 & 4.38 & 0.00 & 10.31 & 18.33 & 0.00 & 0.00 & 0.00 & 0.00 & 13.66 & 0.02 & 20.54 & 0.00 & 0.00 & 0.00 & 1.21 & 0.00 & 0.00 & 6.70 \\ %
        3DSketch*~\cite{3d-sketch} & $x^{\text{rgb}}$,$\hat{x}^{\text{TSDF}}$ & 33.30 & 41.32 & 21.63 & 0.00 & 0.00 & 14.81 & 18.59 & 0.00 & 0.00 & 0.00 & 0.00 & 19.09 & 0.00 & 26.40 & 0.00 & 0.00 & 0.00 & 0.73 & 0.00 & 0.00 & 7.50  \\ %
        AICNet*~\cite{li2020anisotropic} & $x^{\text{rgb}}$,$\hat{x}^{\text{depth}}$ & 29.59 & 43.55 & 20.55 & 11.97 & 0.07 & 12.94 & 14.71 & 4.53 & 0.00 & 0.00 & 0.00 & 15.37 & 2.90 & 28.71 & 0.00 & 0.00 & 0.00 & 2.52 & 0.06 & 0.00 & 8.31  \\ %
        JS3C-Net*~\cite{JS3CNet} & $\hat{x}^{\text{pts}}$ & \textbf{38.98} & 50.49 & 23.74 & 11.94 & 0.07 & \textbf{15.03} & 24.65 & 4.41 & 0.00 & 0.00 & 6.15 & 18.11 & \textbf{4.33} & 26.86 & 0.67 & 0.27 & 0.00 & 3.94 & 3.77 & 1.45 & 10.31  \\
        MonoScene$\dagger$~\cite{cao2022monoscene} & $x^{\text{rgb}}$ & 36.86 & 56.52 & 26.72 & 14.27 & 0.46 & 14.09 & 23.26 & 6.98 & 0.61 & 0.45 & 1.48 & 17.89 & 2.81 & 29.64 & 1.86 & 1.20 & 0.00 & 5.84 & 4.14 & 2.25 & 11.08 \\
        TPVFormer~\cite{tpvformer} & $x^{\text{rgb}}$ & 35.61 & 56.50 & 25.87 & \textbf{20.60} & \textbf{0.85} & 13.88 & 23.81 & 8.08 & 0.36 & 0.05 & 4.35 & 16.92 & 2.26 & 30.38 & 0.51 & 0.89 & 0.00 & \textbf{5.94} & 3.14 & 1.52 & 11.36 \\
        \midrule
        OccFormer & $x^{\text{rgb}}$ & 36.50 & \textbf{58.85} & \textbf{26.88} & 19.61 & 0.31 & 14.40 & \textbf{25.09} & \textbf{25.53} & \textbf{0.81} & \textbf{1.19} & \textbf{8.52} & \textbf{19.63} & 3.93 & \textbf{32.62} & \textbf{2.78} & \textbf{2.82} & 0.00 & 5.61 & \textbf{4.26} & \textbf{2.86} & \textbf{13.46} \\
        \bottomrule
    \end{tabular}\\
    \label{table:kitti_val_perf}
    \vspace{-3mm}
\end{table*}

\subsubsection{Preserve-Pooling}
\label{sec:preserve_pooling}

When converting the high-resolution mask predictions into the low-resolution attention masks for the next iteration, Mask2Former~\cite{mask2former} employs the bilinear interpolation for downsampling. The operation is sufficient to protect the local structures because the image segmentation masks are more complete and contiguous. 
However, we found its trivial adaptation, namely trilinear interpolation, cannot well handle the 3D semantic occupancy prediction. Since the LiDAR-generated segmentation masks for 3D objects are usually partial and sparse, the trilinear downsampling can remove the local structures or even the entire objects. 
To this end, we propose the preserve-pooling by simply using the max-pooling for downsampling the attention masks. Despite a minor modification, we demonstrate its effectiveness in the ablation studies (\cref{sec:ablation}).

\vspace{-3mm}
\subsubsection{Class-Guided Sampling}
\label{sec:class_guided_sampling}

For efficient training, Mask2Former uniformly (or further with importance sampling~\cite{Pointrend}) samples $K$ points in the image space when computing the matching costs and final losses. However, in the 3D occupancy space, the uniform sampling struggles to capture foreground regions, particularly the minor classes, due to sparsity and class imbalance.
To address this issue, we propose the class-guided sampling method. More specifically, we first compute the class frequencies $\mathbf{n}_c \in \mathbb{R}^{N_c}$ from the training set, where $N_c$ is the number of classes. 
Then we compute their reciprocal $\mathbf{w}_c = 1 / \mathbf{n}_c$ and normalize its minimum to 1 with $\mathbf{w}_c = \mathbf{w}_c / \min(\mathbf{w}_c)$. Finally, the sampling weights are computed as $\mathbf{w}_c = \left(\mathbf{w}_c\right)^\beta$,
where $\beta$ is a hyper-parameter.

During training, each voxel is assigned a sampling weight according to its ground-truth class. We then use the multinomial distribution to sample $K$ voxel positions for matching and supervision. 
Note that for nuScenes dataset with only sparse LiDAR point supervisions, we simply use the LiDAR points and random coordinates in a 1:1 ratio as the sampled points. 

\definecolor{nbarrier}{RGB}{255, 120, 50}
\definecolor{nbicycle}{RGB}{255, 192, 203}
\definecolor{nbus}{RGB}{255, 255, 0}
\definecolor{ncar}{RGB}{0, 150, 245}
\definecolor{nconstruct}{RGB}{0, 255, 255}
\definecolor{nmotor}{RGB}{200, 180, 0}
\definecolor{npedestrian}{RGB}{255, 0, 0}
\definecolor{ntraffic}{RGB}{255, 240, 150}
\definecolor{ntrailer}{RGB}{135, 60, 0}
\definecolor{ntruck}{RGB}{160, 32, 240}
\definecolor{ndriveable}{RGB}{255, 0, 255}
\definecolor{nother}{RGB}{139, 137, 137}
\definecolor{nsidewalk}{RGB}{75, 0, 75}
\definecolor{nterrain}{RGB}{150, 240, 80}
\definecolor{nmanmade}{RGB}{213, 213, 213}
\definecolor{nvegetation}{RGB}{0, 175, 0}

\begin{table*}[t]
    \footnotesize
    \setlength{\tabcolsep}{0.0045\linewidth}
    \caption{\textbf{LiDAR segmentation results on nuScenes test set.} The proposed OccFormer outperforms the only vision-based method TPVFormer~\cite{tpvformer} and achieves comparable performance with LiDAR-based methods.}
    \vspace{-2mm}
    \newcommand{\classfreq}[1]{{~\tiny(\nuscenesfreq{#1}\%)}}  %
    \centering
    \begin{tabular}{l|c|c | c c c c c c c c c c c c c c c c}
        \toprule
        Method
        & \makecell{Input \\ Modality} & mIoU
        & \rotatebox{90}{\textcolor{nbarrier}{$\blacksquare$} barrier}
        & \rotatebox{90}{\textcolor{nbicycle}{$\blacksquare$} bicycle}
        & \rotatebox{90}{\textcolor{nbus}{$\blacksquare$} bus}
        & \rotatebox{90}{\textcolor{ncar}{$\blacksquare$} car}
        & \rotatebox{90}{\textcolor{nconstruct}{$\blacksquare$} const. veh.}
        & \rotatebox{90}{\textcolor{nmotor}{$\blacksquare$} motorcycle}
        & \rotatebox{90}{\textcolor{npedestrian}{$\blacksquare$} pedestrian}
        & \rotatebox{90}{\textcolor{ntraffic}{$\blacksquare$} traffic cone}
        & \rotatebox{90}{\textcolor{ntrailer}{$\blacksquare$} trailer}
        & \rotatebox{90}{\textcolor{ntruck}{$\blacksquare$} truck}
        & \rotatebox{90}{\textcolor{ndriveable}{$\blacksquare$} drive. suf.}
        & \rotatebox{90}{\textcolor{nother}{$\blacksquare$} other flat}
        & \rotatebox{90}{\textcolor{nsidewalk}{$\blacksquare$} sidewalk}
        & \rotatebox{90}{\textcolor{nterrain}{$\blacksquare$} terrain}
        & \rotatebox{90}{\textcolor{nmanmade}{$\blacksquare$} manmade}
        & \rotatebox{90}{\textcolor{nvegetation}{$\blacksquare$} vegetation}
        \\
        \midrule
        MINet~\cite{Minet} & LiDAR & 56.3 & 54.6 & 8.2 & 62.1 & 76.6 & 23.0 & 58.7 & 37.6 & 34.9 & 61.5 & 46.9 & 93.3 & 56.4 & 63.8 & 64.8 & 79.3 & 78.3  \\
        
        PolarNet~\cite{zhang2020polarnet} & LiDAR & 69.4 & 72.2 & 16.8 & 77.0 & 86.5 & 51.1 & 69.7 & 64.8 & 54.1 & 69.7 & 63.5 & 96.6 & 67.1 & 77.7 & 72.1 & 87.1 & 84.5  \\
        
        PolarSteam~\cite{chen2021polarstream} & LiDAR & 73.4 & 71.4 & 27.8 & 78.1 & 82.0 & 61.3 & 77.8 & 75.1 & 72.4 & 79.6 & 63.7 & 96.0 & 66.5 & 76.9 & 73.0 & 88.5 & 84.8  \\
        
        JS3C-Net~\cite{JS3CNet} & LiDAR & 73.6	& 80.1	& 26.2 & 87.8 & 84.5 & 55.2	& 72.6	& 71.3	& 66.3	& 76.8	& 71.2	& 96.8	& 64.5	& 76.9	& 74.1	& 87.5	& 86.1 \\
        
        AMVNet~\cite{liong2020amvnet} & LiDAR & 77.3 & 80.6 & 32.0 & 81.7 & 88.9 & 67.1 & 84.3 & 76.1 & 73.5 & 84.9 & 67.3 & 97.5 & 67.4 & 79.4 & 75.5 & 91.5 & 88.7   \\
        
        SPVNAS~\cite{SPVNAS} & LiDAR & 77.4  & 80.0 & 30.0 & 91.9 & 90.8 & 64.7 & 79.0 & 75.6 & 70.9 & 81.0 & 74.6 & 97.4 & 69.2 & 80.0 & 76.1 & 89.3 & 87.1  \\
        
        Cylinder3D++~\cite{cylindrical++} & LiDAR & 77.9 & 82.8 & 33.9 & 84.3 & 89.4 & 69.6 & 79.4 & 77.3 & 73.4 & 84.6 & 69.4 & 97.7 & 70.2 & 80.3 & 75.5 & 90.4 & 87.6   \\
        
        AF2S3Net~\cite{2-s3net} & LiDAR & 78.3 & 78.9 & \textbf{52.2} & 89.9 & 84.2 & \textbf{77.4} & 74.3 & 77.3 & 72.0 & 83.9 & 73.8 & 97.1 & 66.5 & 77.5 & 74.0 & 87.7 & 86.8   \\
        
        DRINet++~\cite{ye2021drinet++} & LiDAR & 80.4 & \textbf{85.5} & 43.2 & 90.5 & \textbf{92.1} & 64.7 & 86.0 & 83.0 & 73.3 & 83.9 & \textbf{75.8} & 97.0 & \textbf{71.0} & \textbf{81.0} & \textbf{77.7} & 91.6 & \textbf{90.2}   \\
        
        LidarMultiNet~\cite{Lidarmultinet} & LiDAR & \textbf{81.4} & 80.4 & 48.4 & \textbf{94.3} & 90.0 & 71.5 & \textbf{87.2} & \textbf{85.2} & \textbf{80.4} & \textbf{86.9} & 74.8 & \textbf{97.8} & 67.3 & 80.7 & 76.5 & \textbf{92.1} & 89.6   \\
        \midrule
        TPVFormer~\cite{tpvformer} & Camera & 69.4  & \textbf{74.0} & 27.5 & 86.3 & 85.5 & \textbf{60.7} & 68.0 & 62.1 & 49.1 & 81.9 & \textbf{68.4} & 94.1 & 59.5 & 66.5 & 63.5 & 83.8 & 79.9  \\ %
        OccFormer (ours) & Camera & \textbf{70.8} & 72.8 &  \textbf{29.9} & \textbf{87.9} & \textbf{85.6} & 57.1 & \textbf{74.9} & \textbf{63.2} & \textbf{53.4} & \textbf{83.0} & 67.6 & \textbf{94.8} & \textbf{61.9} & \textbf{70.0} & \textbf{66.0} & \textbf{84.0} & \textbf{80.5}\\
        \bottomrule
    \end{tabular}
    \label{tab:nusc_lidarseg_test}
    \vspace{-3mm}
\end{table*}

\subsection{Loss Functions}
Following Mask2Former~\cite{maskformer}, we compute the bipartite matching between the predicted and ground-truth segments, considering only the sampled positions. The matching cost includes the class loss and the binary mask loss. With the optimal matching computed by the Hungarian algorithm~\cite{hungarian}, the mask classification loss $\mathcal{L}_{\text{mask-cls}}$ is computed following the matching cost. Besides, the intermediate depth distribution for view transformation is supervised by the projections of LiDAR points, with the binary cross-entropy loss $\mathcal{L}_{\text{depth}}$ following BEVDepth~\cite{bevdepth}. The final training loss is a simple summation: $\mathcal{L} = \mathcal{L}_{\text{mask-cls}} + \mathcal{L}_{\text{depth}}$.
\section{Experiments}
\subsection{Datasets}
The SemanticKITTI dataset~\cite{behley2019semantickitti} is based on the popular KITTI Odometry Benchmark~\cite{kitti} and focuses on the semantic scene understanding with LiDAR points and front cameras. OccFormer is evaluated by its task of semantic scene completion, but with the monocular left camera as input following MonoScene~\cite{cao2022monoscene}. Specifically, the ground-truth semantic occupancy is represented as the $256\times 256\times 32$ voxel grids. Each voxel is 0.2m$\times$0.2m$\times$0.2m large and annotated with 21 semantic classes (19 semantics, 1 free, 1 unknown). Following~\cite{cao2022monoscene, tpvformer}, the 22 sequences are split into 10/1/11 for train/val/test. 

The nuScenes dataset~\cite{nuscenes} is a large-scale autonomous driving dataset, collected in Boston and Singapore. The dataset includes 1000 driving sequences from various scenes. Each sequence lasts for around 20 seconds and the key-frames are annotated at 2Hz with 3D bounding boxes. The Panoptic nuScenes dataset~\cite{panoptic_nusc} further extends the nuScenes dataset to provide the annotations for LiDAR semantic segmentation. Similar to TPVFormer~\cite{tpvformer}, we train OccFormer with sparse LiDAR point supervisions for 3D semantic occupancy prediction. We follow the official protocol to split the total scenes into train/val/test splits with 700/150/150 scenes. We report quantitative results for the LiDAR segmentation and qualitative visualizations for the 3D semantic occupancy prediction. 

\subsection{Implementation Details}
\paragraph{Network Structures.}
Considering the image backbone network, we adopt EfficientNetB7~\cite{cao2022monoscene} on SemanticKITTI and ResNet-101~\cite{resnet} on nuScenes, following the compared methods~\cite{cao2022monoscene, tpvformer}. The view transformer creates the 3D feature volume of size 128$\times$128$\times$16, with 128 channels. The transformer encoder consists of 4 stages with 2 dual-path transformer blocks each. The generated multi-scale 3D features are projected to 192 channels and processed the multi-scale deformable self-attention with 6 layers. The transformer decoder mainly follows the implementation from Mask2Former~\cite{mask2former}. We increase the number of sampling points to 50176 (4$\times$) and set $\beta$ as 0.25 for the class-guided sampling. The predicted occupancy is upsampled 2$\times$ to 256$\times$256$\times$32 for full-scale evaluation. 

\paragraph{Training Setup.}
Unless specified, we train the model for 30 epochs on SemanticKITTI dataset and 24 epochs on nuScenes dataset. The AdamW~\cite{adamw} optimizer with initial learning rate 1e-4 and weight decay 0.01 is used. The learning rate is decayed by a multi-step scheduler. All models are trained with a batch size of 8 on 8 RTX 3090 GPUs with 24G memory. 
For data augmentation, we use random resize, rotation, and flip for the image space and 3D flip for the 3D volume space, following recent practices for BEV-based 3D object detection~\cite{bevdet, bevdepth, zhang2022beverse}.

\begin{table}
    \footnotesize
    \centering
    \setlength{\tabcolsep}{0.01\linewidth}
    \caption{Ablation study on the dual-path encoder. }
    \vspace{-2mm}
    \begin{tabular}[b]{cc|cc|cc}
        \toprule
        Local & Global & Params & GFLOPs & IoU$\uparrow$ & mIoU$\uparrow$\\
        \midrule
        \cmark & & 74.1M & 494.2 & 36.42 & 12.95 \\
         & \cmark & 81.4M & 407.4 & 36.37 & 12.93 \\
        \cmark & \cmark & 81.4M & 515.3 & \textbf{36.50} & \textbf{13.46} \\
        \midrule
        \multicolumn{2}{c|}{3D ResNet-16~\cite{resnet}} & 132.5M & 825.8 & 36.12 & 12.89 \\
        \multicolumn{2}{c|}{3D Swin-T~\cite{video_swin}} & 82.3M & 437.9 & 36.32 & 12.80 \\
        \bottomrule
    \end{tabular}
    \label{tab:ablation_encoder}
    \vspace{-2mm}
\end{table}


\begin{figure*}
    \centering
    \newcolumntype{P}[1]{>{\centering\arraybackslash}m{#1}}
    \setlength{\tabcolsep}{0.001\textwidth}
    \renewcommand{\arraystretch}{0.8}
    \footnotesize
    \begin{tabular}{P{0.22\textwidth} P{0.22\textwidth} P{0.22\textwidth} P{0.22\textwidth}}		
    Input & MonoScene~\cite{cao2022monoscene} & OccFormer (ours) & Ground Truth
    \\[-0.1em]
    \includegraphics[width=.7\linewidth]{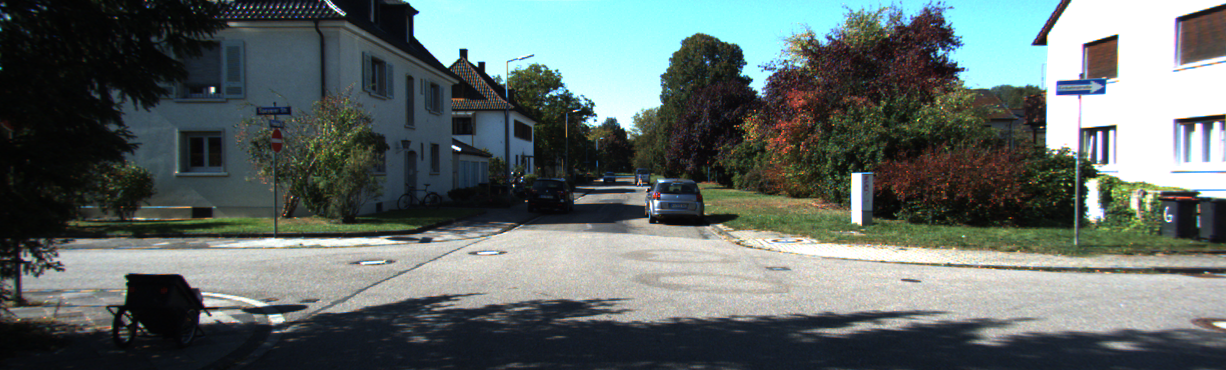} &  
    \includegraphics[width=.7\linewidth]{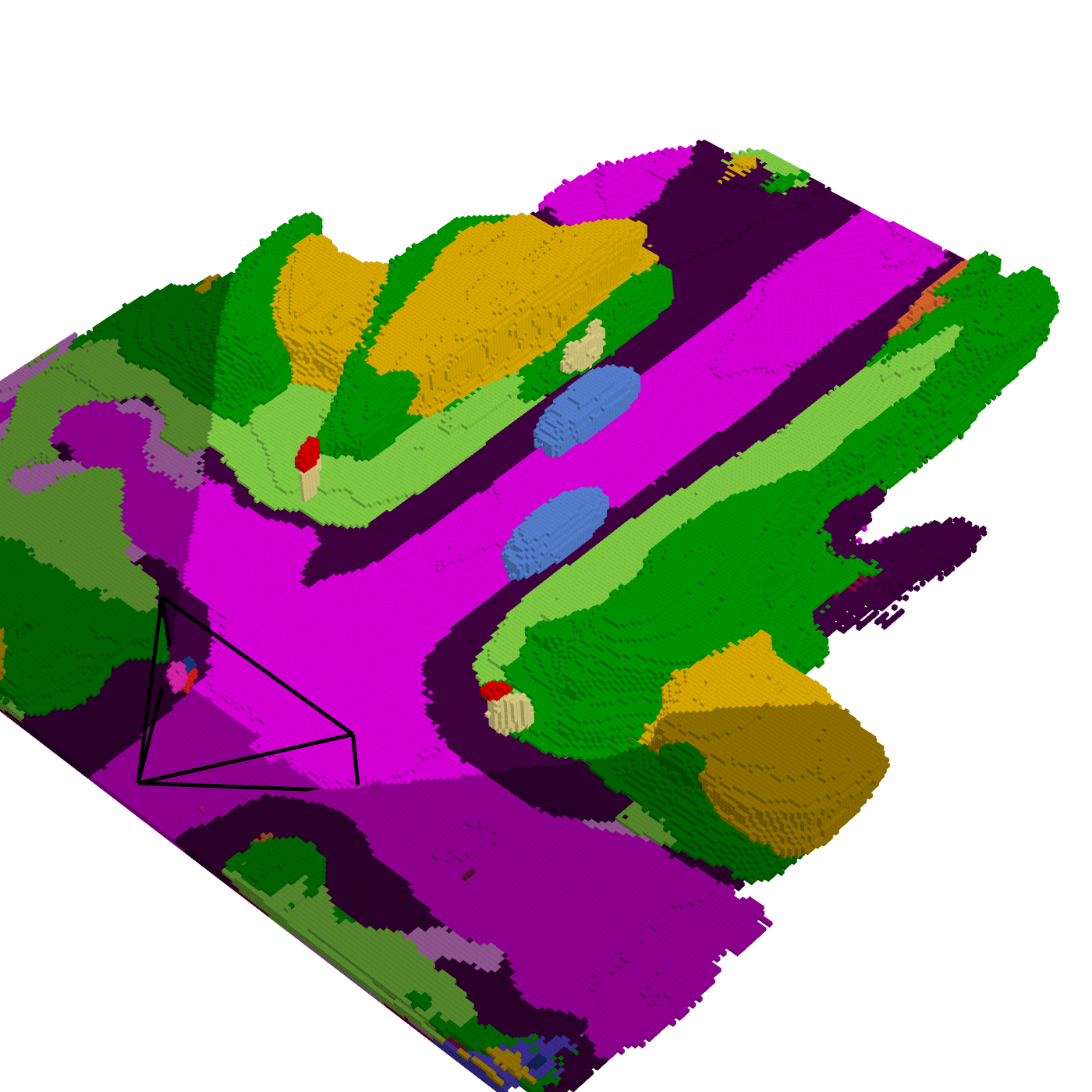} & 
    \includegraphics[width=.7\linewidth]{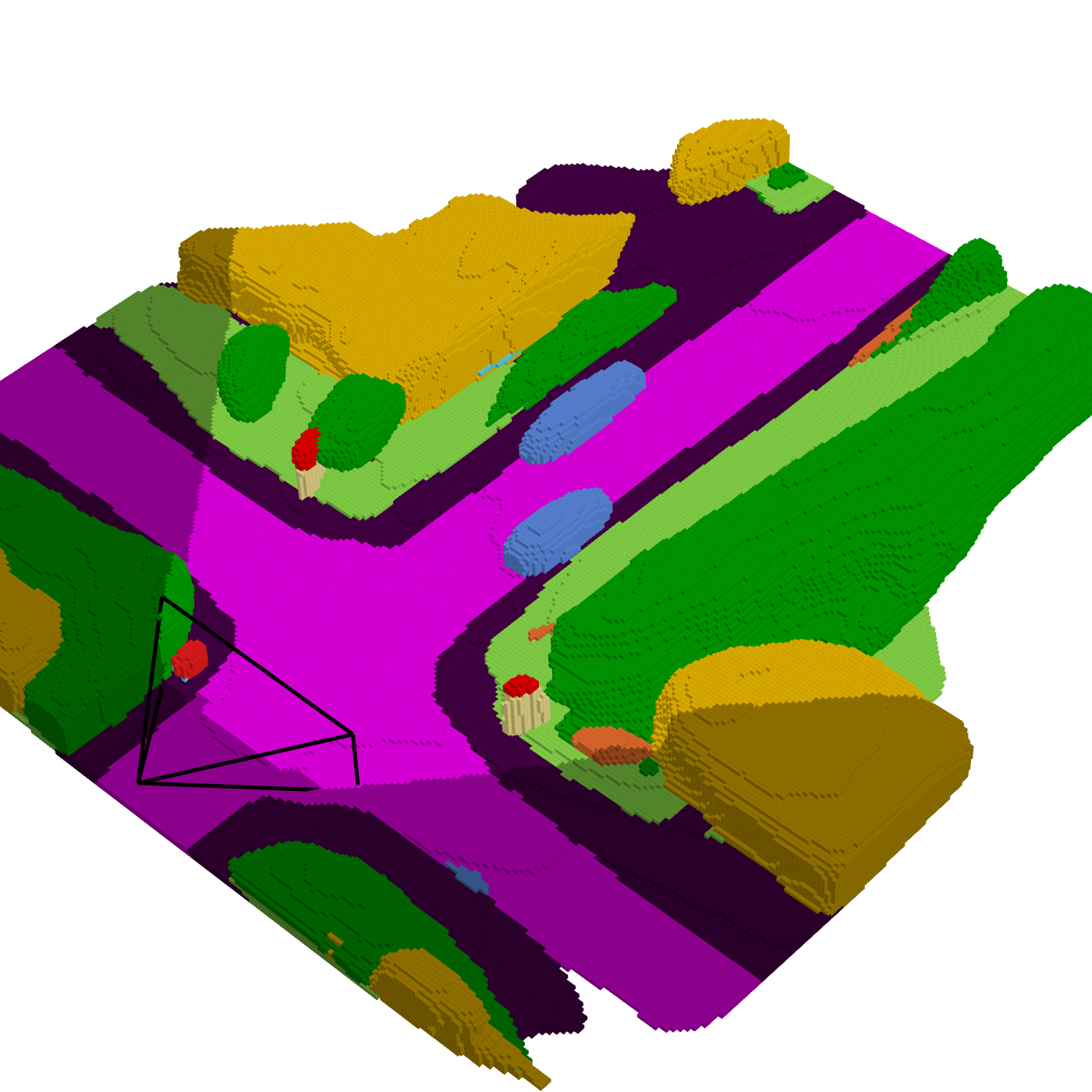} & 
    \includegraphics[width=.7\linewidth]{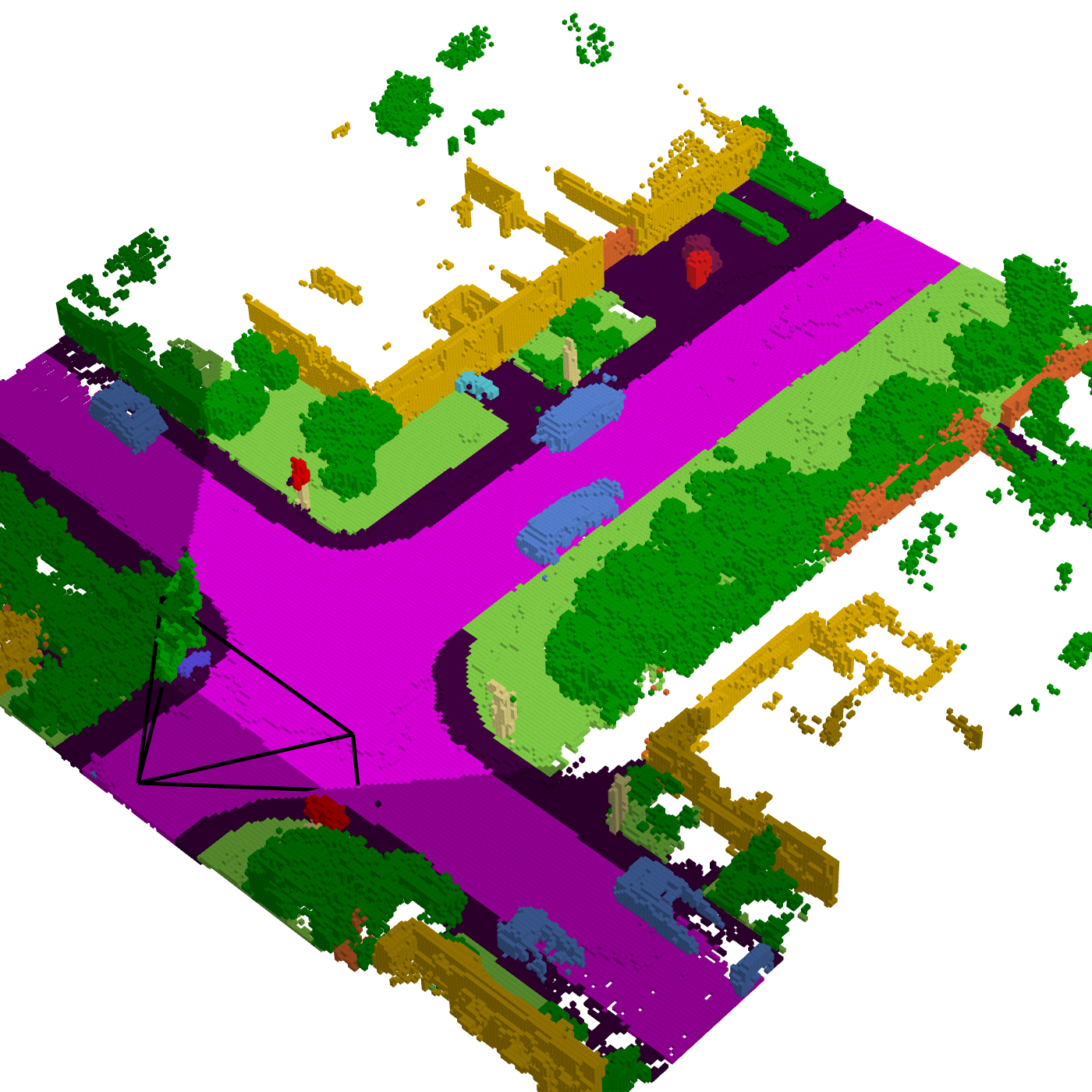}
    \\[-0.8em]
    \includegraphics[width=.7\linewidth]{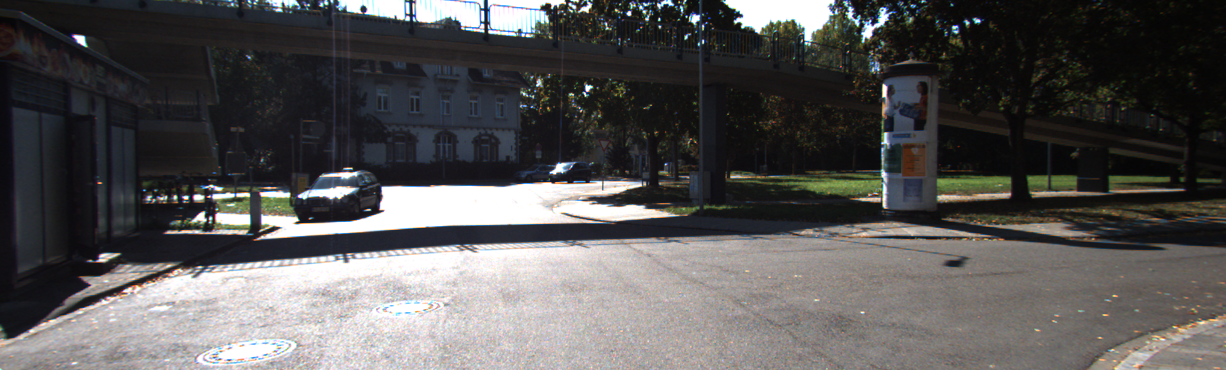} &  
    \includegraphics[width=.7\linewidth]{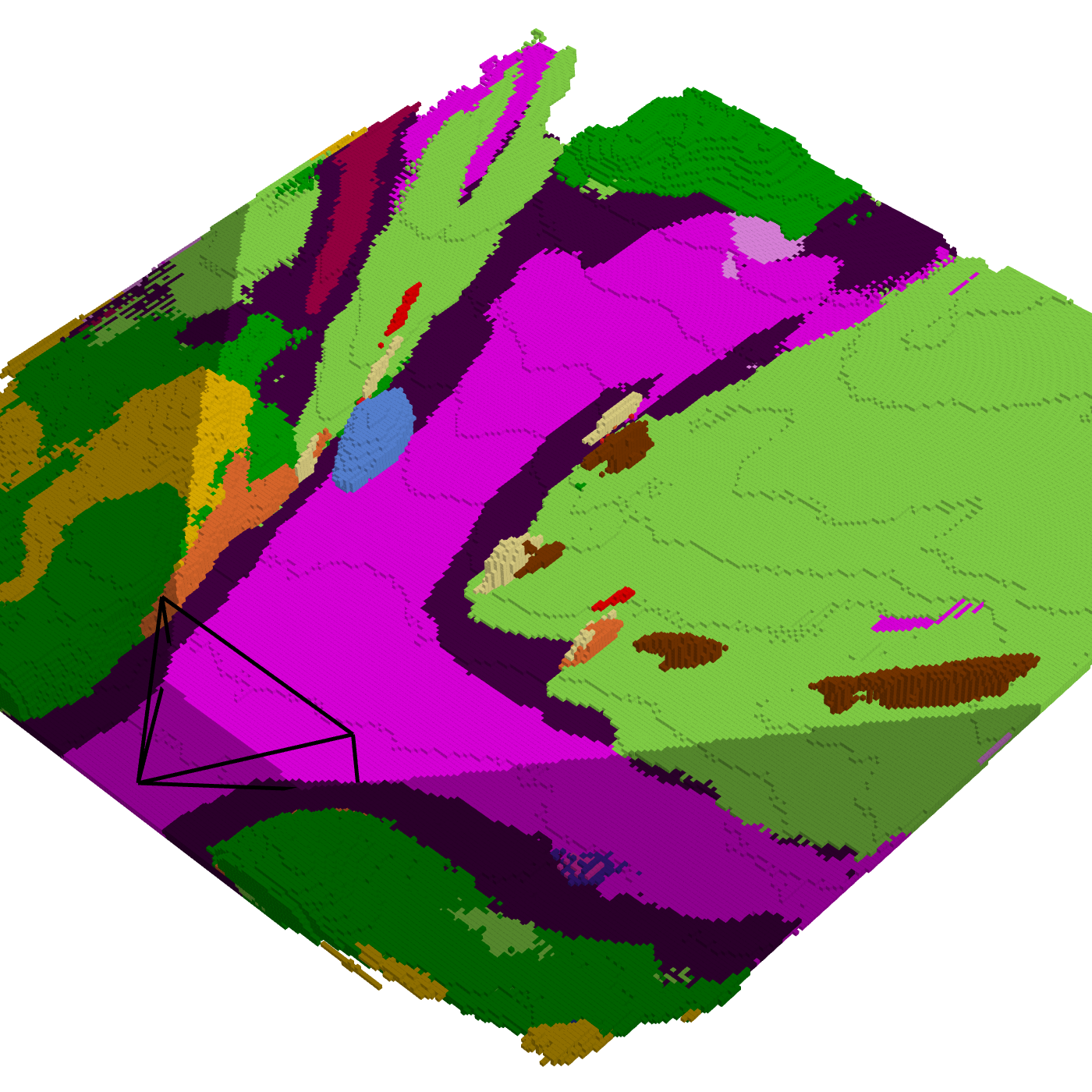} & 
    \includegraphics[width=.7\linewidth]{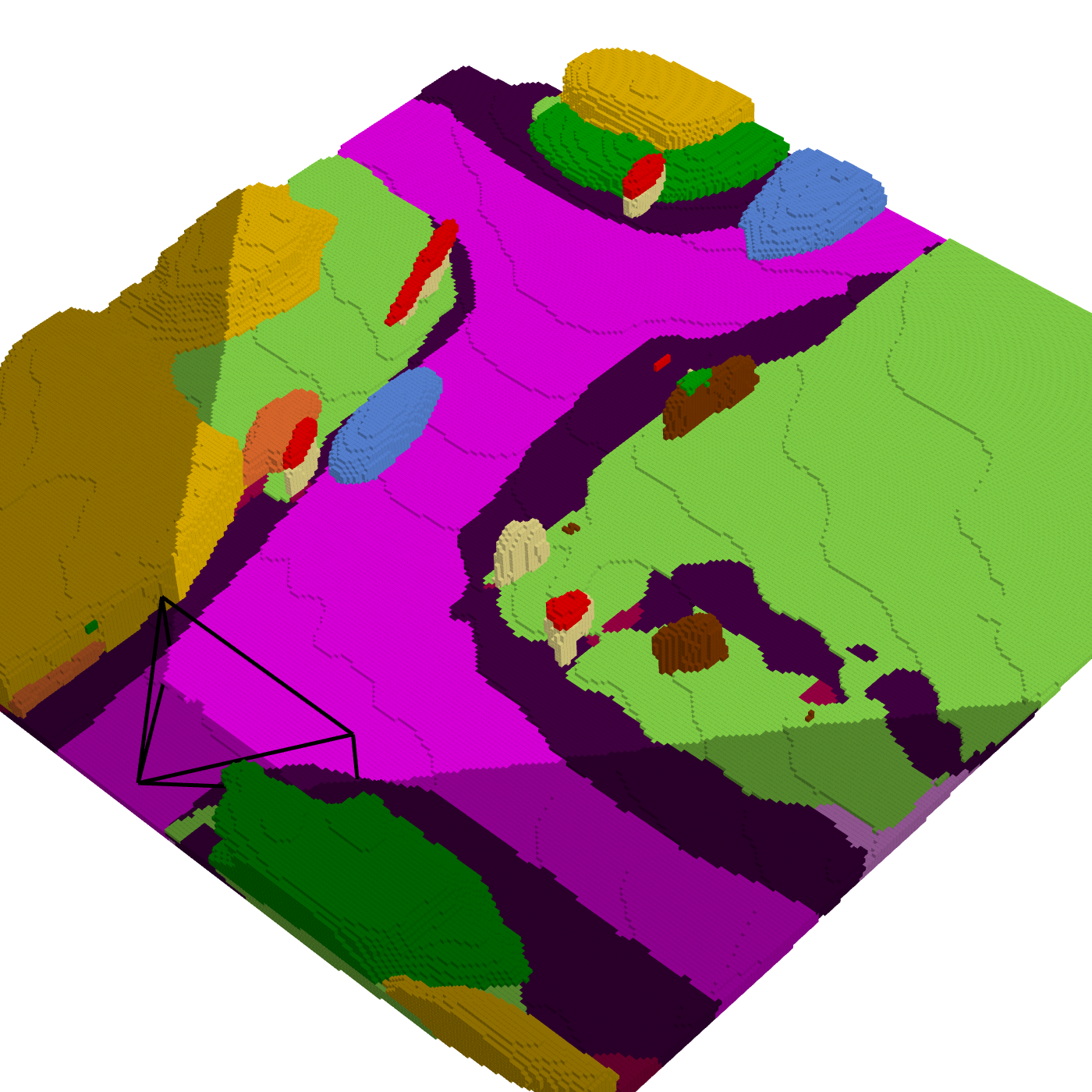} & 
    \includegraphics[width=.7\linewidth]{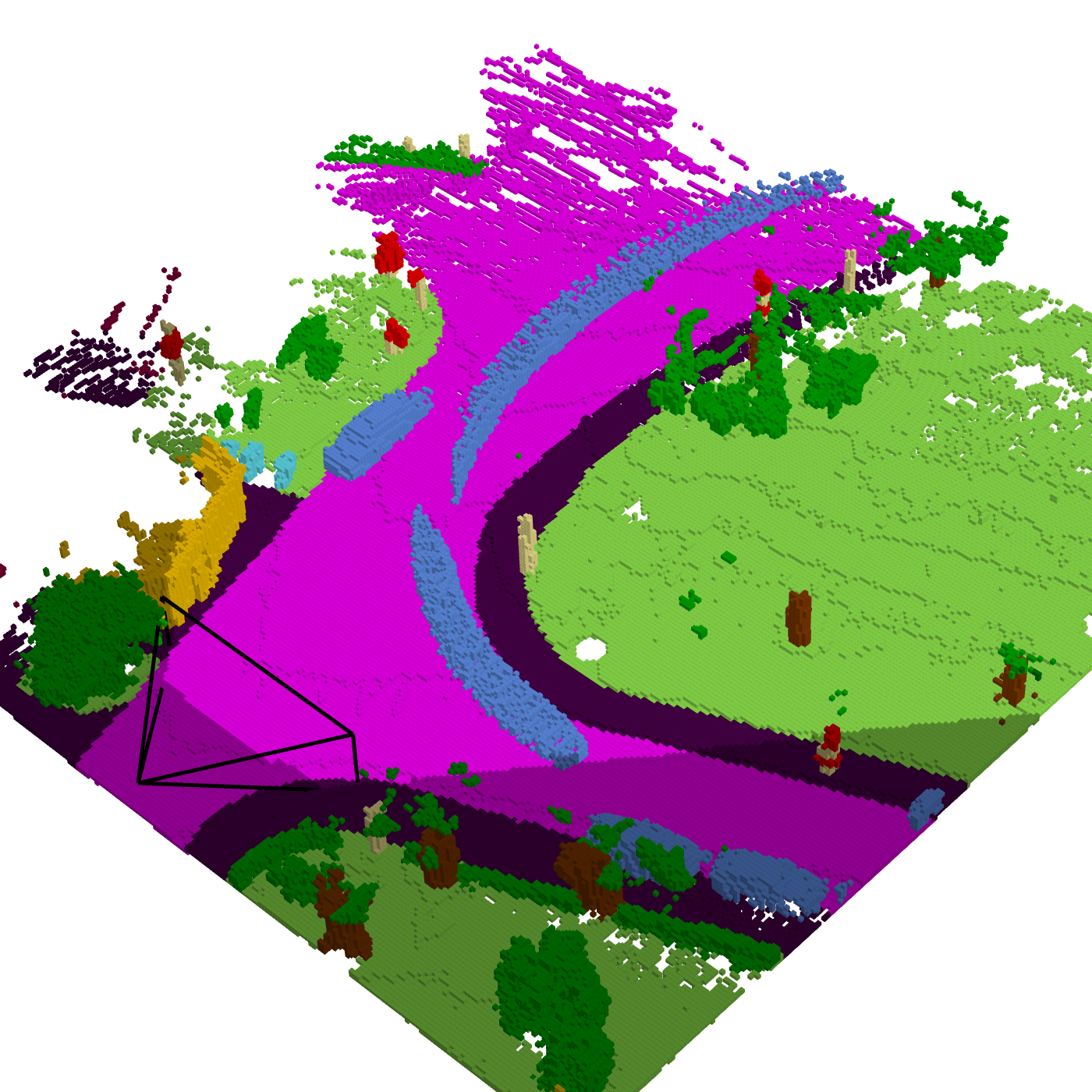} \\
    \\[-0.8em]
    \includegraphics[width=.7\linewidth]{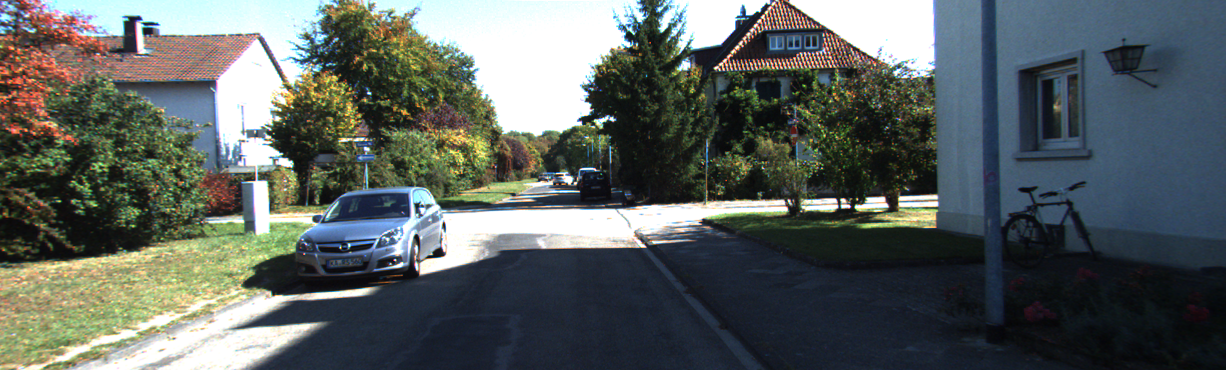} &  
    \includegraphics[width=.7\linewidth]{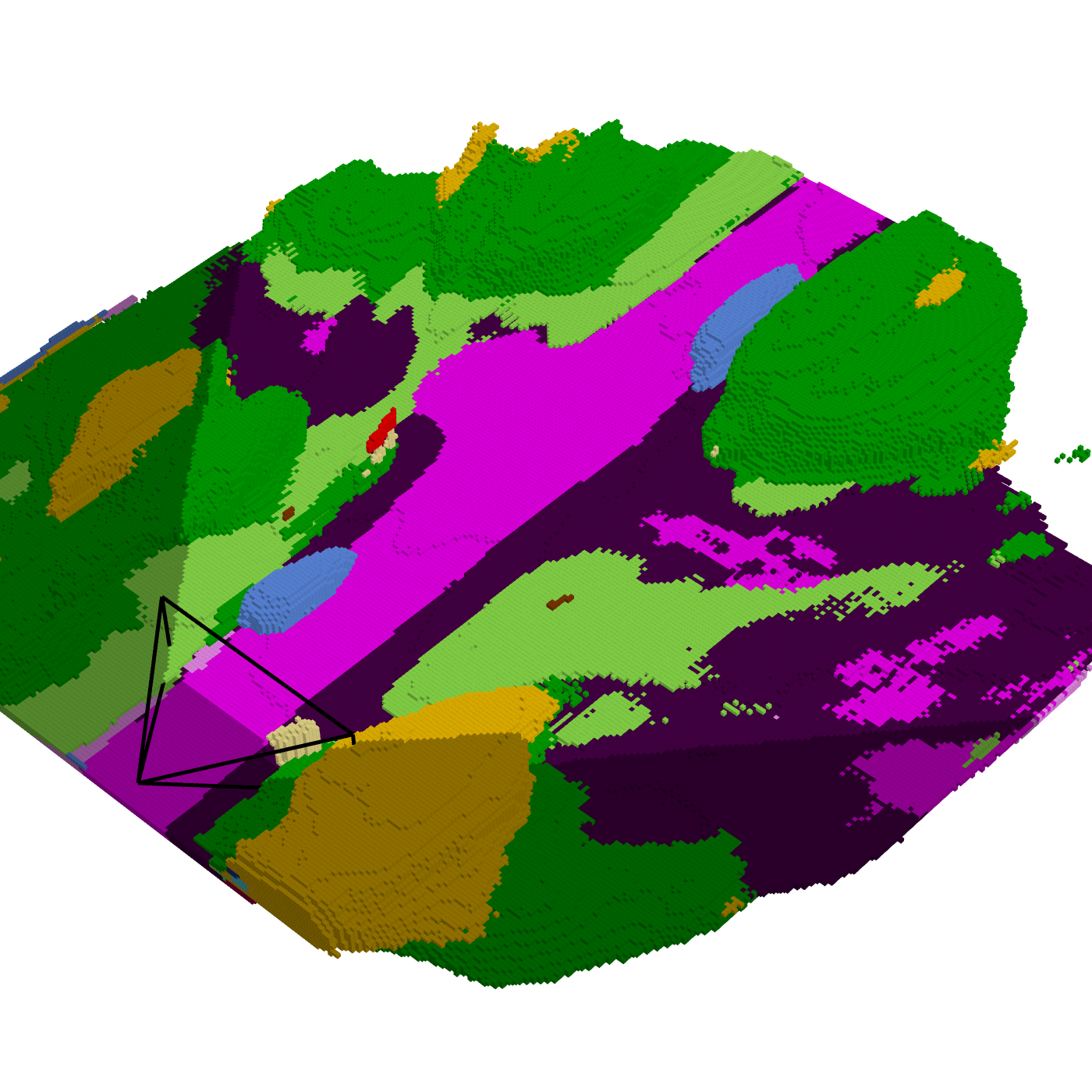} & 
    \includegraphics[width=.7\linewidth]{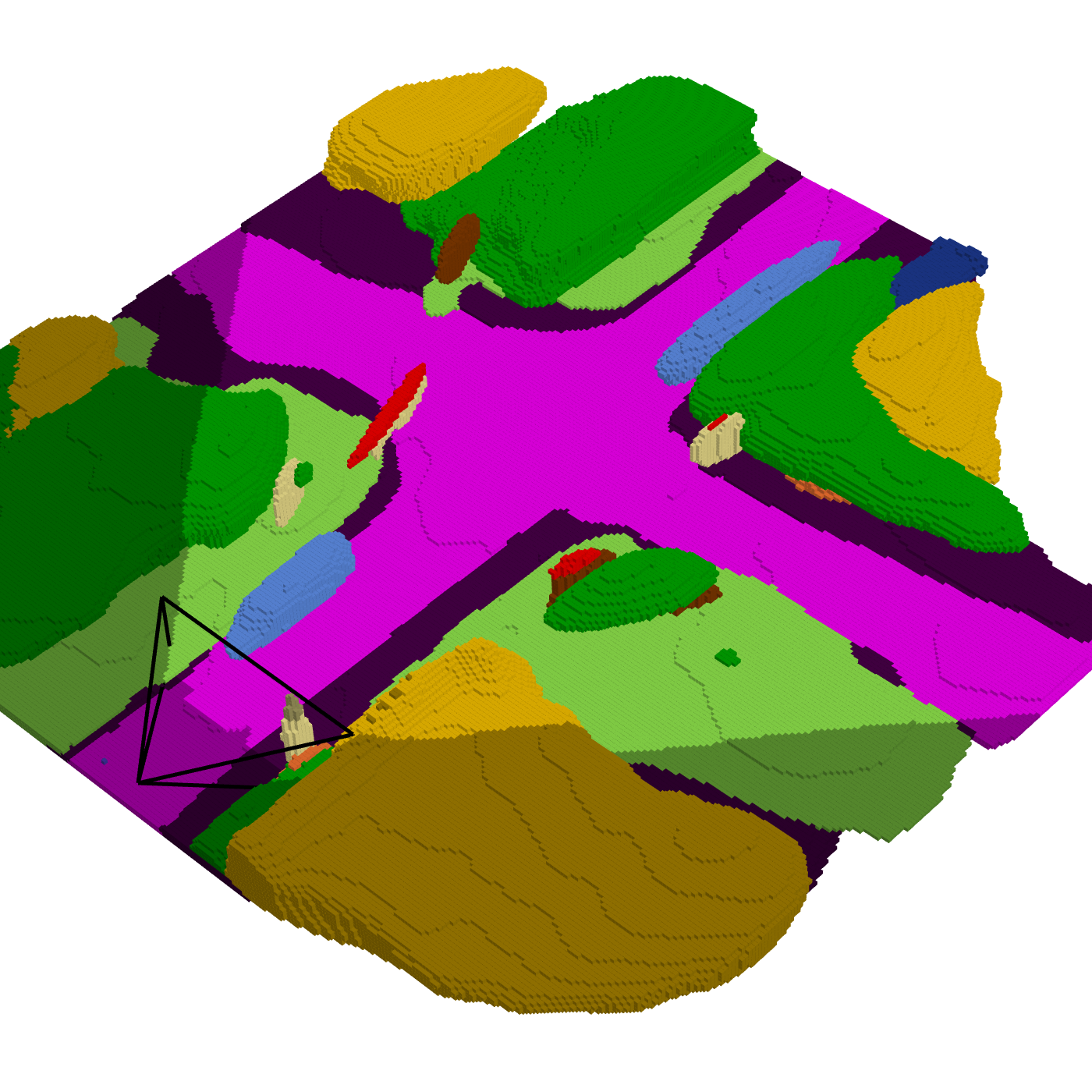} & 
    \includegraphics[width=.7\linewidth]{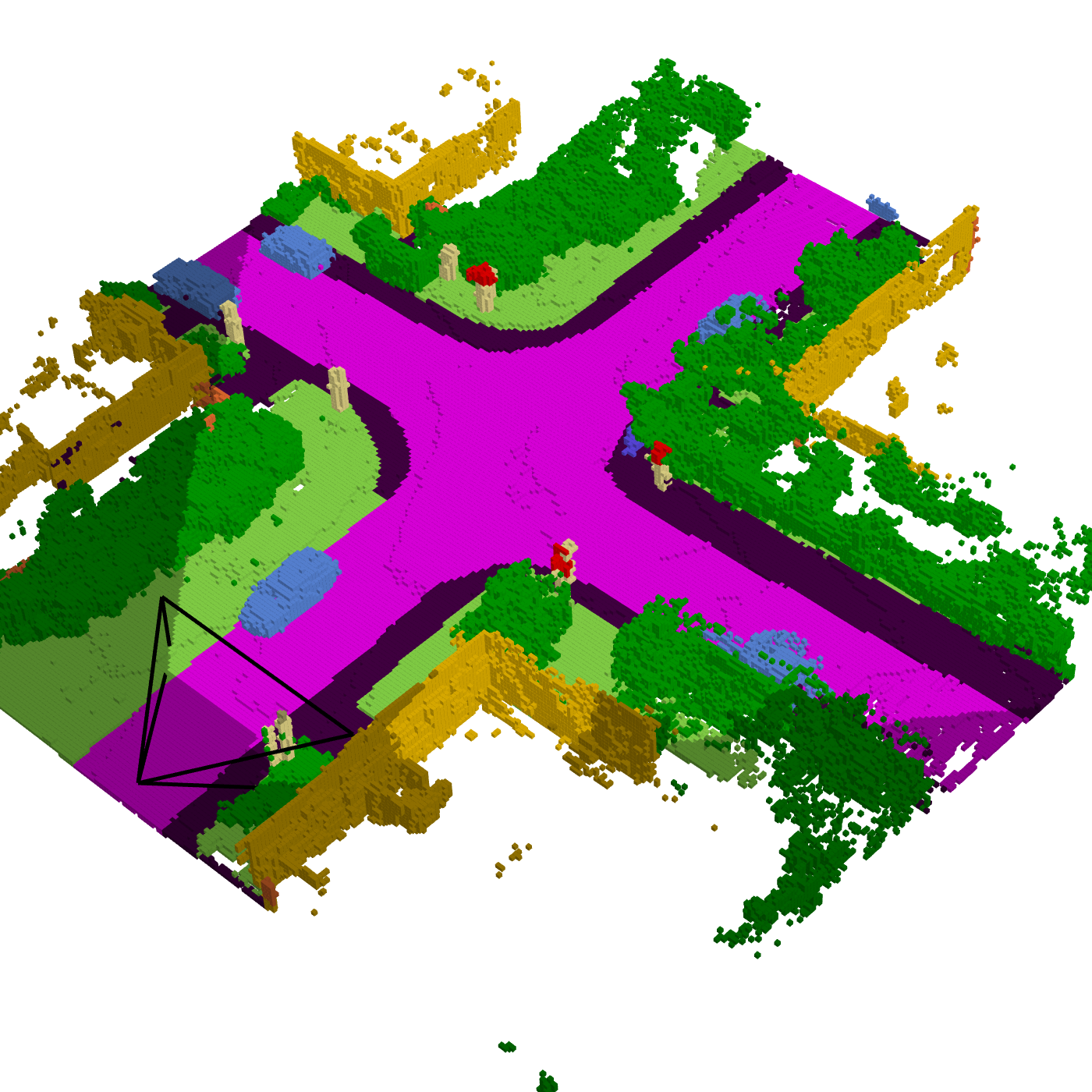} \\
    \multicolumn{4}{c}{
    \scriptsize
    \textcolor{bicycle}{$\blacksquare$}bicycle~
    \textcolor{car}{$\blacksquare$}car~
    \textcolor{motorcycle}{$\blacksquare$}motorcycle~
    \textcolor{truck}{$\blacksquare$}truck~
    \textcolor{other-vehicle}{$\blacksquare$}other vehicle~
    \textcolor{person}{$\blacksquare$}person~
    \textcolor{bicyclist}{$\blacksquare$}bicyclist~
    \textcolor{motorcyclist}{$\blacksquare$}motorcyclist~
    \textcolor{road}{$\blacksquare$}road~
    \textcolor{parking}{$\blacksquare$}parking~}\\
    \multicolumn{4}{c}{
    \scriptsize
    \textcolor{sidewalk}{$\blacksquare$}sidewalk~
    \textcolor{other-ground}{$\blacksquare$}other ground~
    \textcolor{building}{$\blacksquare$}building~
    \textcolor{fence}{$\blacksquare$}fence~
    \textcolor{vegetation}{$\blacksquare$}vegetation~
    \textcolor{trunk}{$\blacksquare$}trunk~
    \textcolor{terrain}{$\blacksquare$}terrain~
    \textcolor{pole}{$\blacksquare$}pole~
    \textcolor{traffic-sign}{$\blacksquare$}traffic sign			
    }
    \end{tabular}
    \vspace{-1mm}
    \caption{\textbf{Qualitative results on SemanticKITTI validation set.} The input monocular image is shown on the left and the 3D semantic occupancy results from MonoScene~\cite{cao2022monoscene}, our OccFormer, and the annotations are then visualized sequentially. The darker colors within the occupancy images represent the unseen parts out of the camera FOV.}
    \label{fig:qualitative_kitti}
    \vspace{-4mm}
\end{figure*} 

\subsection{Metrics}
We report the mean intersection over union (mIoU) for both the semantic scene completion (SSC) and the LiDAR segmentation tasks. Also, the intersection over union (IoU) for the class-agnostic scene completion (SC) task is reported. To infer the LiDAR segmentation results, the LiDAR points are only used to query their corresponding semantic logits from the predicted 3D semantic occupancy volume.

\subsection{Main Results}
\begin{table}
    \footnotesize
    \centering
    \setlength{\tabcolsep}{0.01\linewidth}
    \caption{Ablation study on the pixel decoder. }
    \vspace{-2mm}
    \begin{tabular}[b]{cc|cc|cc}
        \toprule
        Method & Layer & params & GFLOPs & IoU$\uparrow$ & mIoU$\uparrow$\\
        \midrule
        MsDeAttn3D & 3 & 2.74M & 329.3 & 35.74 & 13.22\\
        MsDeAttn3D & 6 & 4.07M & 379.2 & \textbf{36.50} & \textbf{13.46} \\
        \midrule
        FPN-3D~\cite{FPN} & - & 4.35M & 307.0 & 36.12 & 12.89 \\
        \bottomrule
    \end{tabular}
    \label{tab:ablation_neck3d}
    \vspace{-4mm}
\end{table}

\paragraph{Semantic Scene Completion.}
As shown in~\cref{table:kitti_test_perf}, we report the quantitative comparison of existing monocular methods for the semantic scene completion task on SemanticKITTI test set. 
We can observe that OccFormer outperforms all existing competitors, especially for the more challenging task of semantic scene completion. Compared with the recent TPVFormer~\cite{tpvformer}, our method achieves a remarkable boost of 1.06 mIoU, demonstrating the effectiveness of OccFormer for semantic scene completion. Also, we report the results on SemanticKITTI validation set in~\cref{table:kitti_val_perf}. OccFormer achieves comparable IoU for scene completion and significantly better performance for the SSC mIoU. 

\begin{figure*}[t]
\centering
\includegraphics[width=1.0\linewidth]{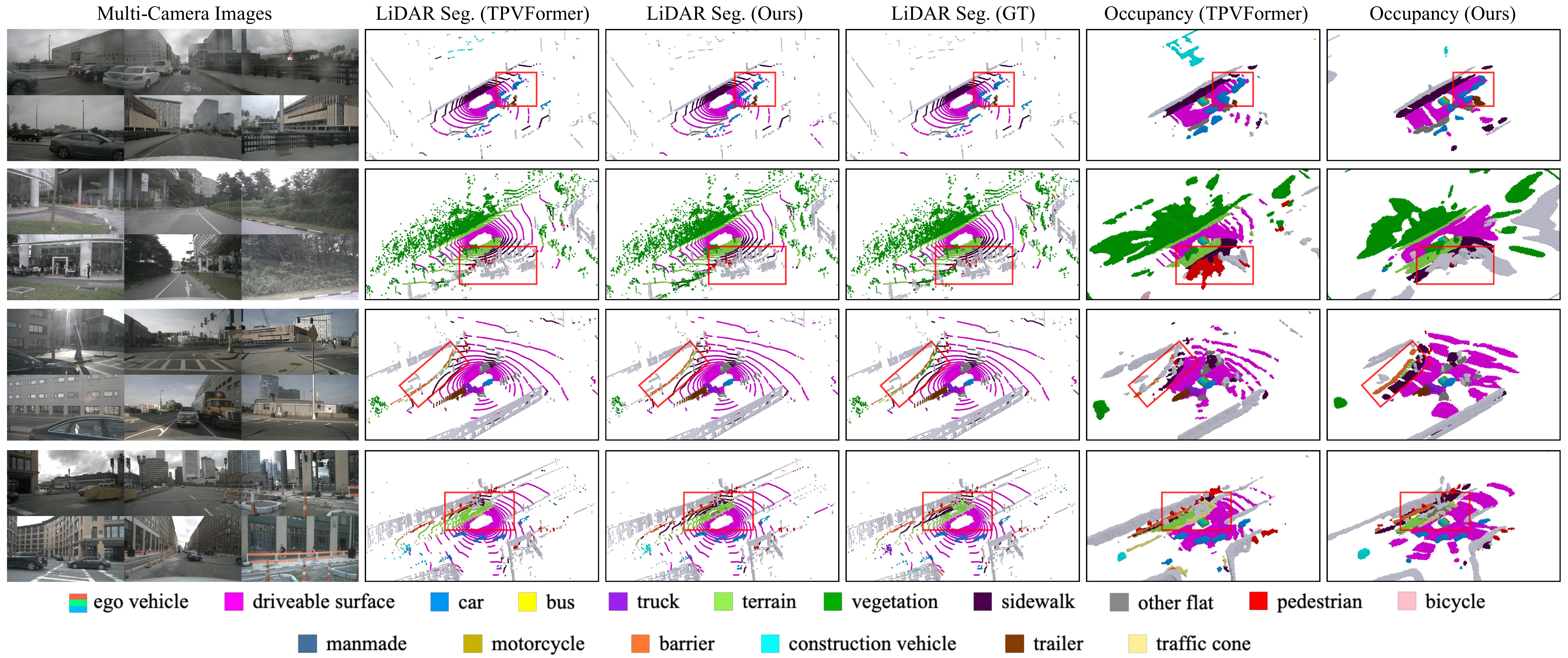}
\vspace{-2mm}
\caption{
\textbf{Qualitative results on nuScenes validation set.} The leftmost column shows the input surrounding images, the following three columns visualize the LiDAR segmentation from TPVFormer~\cite{tpvformer}, our method, and the annotation. The final two columns visualize the predicted 3D semantic occupancy from TPVFormer and our method.}
\label{fig:nusc_visualize_compare}
\vspace{-3mm}
\end{figure*}

\vspace{-3mm}
\paragraph{LiDAR Semantic Segmentation.}
Following the practices from TPVFormer~\cite{tpvformer}, the LiDAR semantic segmentation task is utilized as a quantitative indicator for the 3D semantic occupancy prediction. As shown in~\cref{tab:nusc_lidarseg_test}, our method outperforms the only vision-based method TPVFormer and achieves comparable performance with the state-of-the-art LiDAR-based methods. Note that our method requires only one model to perform both the LiDAR segmentation and the semantic occupancy prediction, while the TPVFormer~\cite{tpvformer} model trained for LiDAR segmentation cannot produce reasonable occupancy predictions. The results on nuScenes validation set is included in~\cref{sec:appendix_nusc_lidarseg_val}.

\subsection{Ablation Studies}
\label{sec:ablation}

The ablation is conducted on SemanticKITTI validation set and from three perspectives: the dual-path encoder, the pixel decoder, and the transformer decoder.

\begin{table}
    \footnotesize
    \centering
    \setlength{\tabcolsep}{0.01\linewidth}
    \caption{Ablation study on the transformer decoder. }
    \vspace{-2mm}
    \begin{tabular}[b]{cc|cc}
        \toprule
        Resize method & Sampling method & IoU$ \uparrow$ & mIoU$ \uparrow$\\
        \midrule
        Tri-linear & Uniform & 35.04 & 11.61 \\
        Max-pool & Uniform & 35.41 & 12.13 \\
        Tri-linear & Class-guided & 36.21 & 13.01 \\
        Max-pool & Class-guided & \textbf{36.50} & \textbf{13.46} \\
        \bottomrule
    \end{tabular}
    \vspace{-3mm}
    \label{tab:ablation_mask2former}
\end{table}

\vspace{-2mm}
\paragraph{Ablation on the Dual-path Encoder.}
In~\cref{tab:ablation_encoder}, we ablate the dual-path design for the 3D feature extraction and compare it with other baseline methods. First, both the local and global paths contribute to the final performance positively. Since the local and global pathways focus on the fine-grained structures and the scene-level semantic layouts respectively, their complementary influence is quite understandable. Also, our dual-path transformer encoder achieves a better trade-off than the vanilla 3D convolution and the 3D windowed attention proposed in~\cite{video_swin}. 

\vspace{-2mm}
\paragraph{Ablation on the Pixel Decoder.}
In~\cref{tab:ablation_neck3d}, we compare different structures for the pixel decoder, which aims to fuse multi-scale features and generate the per-voxel mask embeddings. Thanks to the dynamic receptive field and multi-scale aggregation, the multi-scale 3D deformable attention performs better than the classic FPN~\cite{FPN}, tailored for 3D. Therefore, we utilize the 6-layer multi-scale 3D deformable attention as the pixel decoder for OccFormer. 

\vspace{-3mm}
\paragraph{Ablation on the Transformer Decoder.}
In~\cref{tab:ablation_mask2former}, we ablate the methods of resizing attention masks and sampling points for supervision. Despite the state-of-the-art performance for 2D segmentation, the naive adaptation of Mask2Former~\cite{mask2former} for 3D semantic occupancy prediction achieves inferior performance, only 11.61 mIoU. 
Compared with the tri-linear interpolation, we employ the max-pooling to preserve the fine-grained 3D predictions during downsampling, which achieves a boost of about 0.5 mIoU.
On the other hand, the proposed class-guided sampling significantly outperforms the default uniform sampling because it can better adapt to the task of 3D semantic occupancy prediction, with a lot more ``pixels'' but much sparser supervisions than the 2D counterpart. 


\subsection{Qualitative Results}

\vspace{-1mm}
\paragraph{Semantic Scene Completion.}
In~\cref{fig:qualitative_kitti}, we visualize the predicted results of semantic scene completion on SemanticKITTI validation set from MonoScene~\cite{cao2022monoscene} and our proposed OccFormer. Compared with MonoScene, our method can better understand the scene-level semantic layout and hallucinate the invisible regions. Also, OccFormer is good at recovering the object structures and reasoning about the interactions among neighbouring semantic classes. For example, the predicted buildings (in golden yellow) are more complete and located properly with the surrounding vegetation (in dark green), while MonoScene can generate the entangled results.

\vspace{-3mm}
\paragraph{LiDAR Segmentation and 3D Semantic Occupancy.}
We visualize the predictions for LiDAR segmentation and 3D semantic occupancy in~\cref{fig:nusc_visualize_compare}. Note that TPVFormer generates the required outputs with two separately trained models, while our method uses one single model. Nonetheless, OccFormer still achieves more accurate results on LiDAR segmentation. More importantly, the predicted 3D semantic occupancy from OccFormer is more contiguous, complete, and realistic than TPVFormer. For example, the predicted driveable surface is more contiguous and the foreground objects like cars and traffic cones have more accurate structures. 

\vspace{-2mm}
\section{Conclusion}
\vspace{-1mm}
In this paper, we have presented OccFormer, a dual-path transformer network for camera-based 3D semantic occupancy prediction. 
To effectively process the camera-generated 3D voxel features, we have proposed the dual-path transformer block, which efficiently captures the fine-grained details and scene-level layouts with the local and global pathways. 
Also, we have been the first to employ mask classification models for 3D semantic occupancy prediction. Given the inherent sparsity and class imbalance, the proposed preserve-pooling and class-guided sampling have significantly improved the performance. 
OccFormer has achieved state-of-the-art performance for semantic scene completion on SemanticKITTI test set and for camera-based LiDAR segmentation on nuScenes test set.

\appendix
\appendix

\definecolor{nbarrier}{RGB}{255, 120, 50}
\definecolor{nbicycle}{RGB}{255, 192, 203}
\definecolor{nbus}{RGB}{255, 255, 0}
\definecolor{ncar}{RGB}{0, 150, 245}
\definecolor{nconstruct}{RGB}{0, 255, 255}
\definecolor{nmotor}{RGB}{200, 180, 0}
\definecolor{npedestrian}{RGB}{255, 0, 0}
\definecolor{ntraffic}{RGB}{255, 240, 150}
\definecolor{ntrailer}{RGB}{135, 60, 0}
\definecolor{ntruck}{RGB}{160, 32, 240}
\definecolor{ndriveable}{RGB}{255, 0, 255}
\definecolor{nother}{RGB}{139, 137, 137}
\definecolor{nsidewalk}{RGB}{75, 0, 75}
\definecolor{nterrain}{RGB}{150, 240, 80}
\definecolor{nmanmade}{RGB}{213, 213, 213}
\definecolor{nvegetation}{RGB}{0, 175, 0}

\begin{table*}[ht]
    \footnotesize
    \setlength{\tabcolsep}{0.0045\linewidth}
    \caption{\textbf{LiDAR segmentation results on nuScenes validation set.} For camera-based methods, we list the utilized backbone networks and the input image sizes. OccFormer notably surpasses the recently proposed TPVFormer~\cite{tpvformer} and first achieves 70\%+ mIoU with only multi-view images.}
    \newcommand{\classfreq}[1]{{~\tiny(\nuscenesfreq{#1}\%)}}  %
    \centering
    \begin{tabular}{l|c|c|c|c| c c c c c c c c c c c c c c c c}
    \toprule
    Method
    & \makecell{Input \\ Modality} & Backbone & \makecell{Image \\ Size} & mIoU
    & \rotatebox{90}{\textcolor{nbarrier}{$\blacksquare$} barrier}
    
    & \rotatebox{90}{\textcolor{nbicycle}{$\blacksquare$} bicycle}
    
    & \rotatebox{90}{\textcolor{nbus}{$\blacksquare$} bus}
    
    & \rotatebox{90}{\textcolor{ncar}{$\blacksquare$} car}
    
    & \rotatebox{90}{\textcolor{nconstruct}{$\blacksquare$} const. veh.}
    
    & \rotatebox{90}{\textcolor{nmotor}{$\blacksquare$} motorcycle}
    
    & \rotatebox{90}{\textcolor{npedestrian}{$\blacksquare$} pedestrian}
    
    & \rotatebox{90}{\textcolor{ntraffic}{$\blacksquare$} traffic cone}
    
    & \rotatebox{90}{\textcolor{ntrailer}{$\blacksquare$} trailer}
    
    & \rotatebox{90}{\textcolor{ntruck}{$\blacksquare$} truck}
    
    & \rotatebox{90}{\textcolor{ndriveable}{$\blacksquare$} drive. suf.}
    
    & \rotatebox{90}{\textcolor{nother}{$\blacksquare$} other flat}
    
    & \rotatebox{90}{\textcolor{nsidewalk}{$\blacksquare$} sidewalk}
    
    & \rotatebox{90}{\textcolor{nterrain}{$\blacksquare$} terrain}
    
    & \rotatebox{90}{\textcolor{nmanmade}{$\blacksquare$} manmade}
    
    & \rotatebox{90}{\textcolor{nvegetation}{$\blacksquare$} vegetation}
    
    \\
    \midrule
    
    RangeNet++~\cite{rangenet++} & LiDAR & \multirow{4}{*}{-} & \multirow{4}{*}{-} &  65.5 & 66.0 & 21.3 & 77.2 & 80.9 & 30.2 & 66.8 & 69.6 &  52.1 & 54.2 & {72.3} & {94.1} & 66.6 & 63.5 & 70.1 & 83.1 & 79.8 \\
    
    PolarNet~\cite{zhang2020polarnet} & LiDAR &  &  & 71.0 & 74.7 & 28.2 & 85.3 & 90.9 & 35.1 & 77.5 & 71.3 & 58.8 & 57.4 & 76.1 & 96.5 & 71.1 & 74.7 & 74.0 & 87.3 & 85.7  \\
    
    Salsanext~\cite{Salsanext} & LiDAR & & & 72.2 & 74.8 & 34.1 & 85.9 & 88.4 & 42.2 & 72.4 & 72.2 & 63.1 & 61.3 & 76.5 & 96.0 & 70.8 & 71.2 & 71.5 & 86.7 & 84.4 \\
    
    Cylinder3D++~\cite{cylindrical++} & LiDAR & & & \bf{76.1} & \bf{76.4} & \bf{40.3} & \bf{91.2} & \bf{93.8} & \textbf{51.3} & \bf{78.0} & \bf{78.9} & \bf{64.9} & \bf{62.1} & \bf{84.4} & \bf{96.8} & \bf{71.6} & \bf{76.4} & \bf{75.4} & \bf{90.5} & \bf{87.4}  \\
    \midrule	
    
    TPVFormer~\cite{tpvformer} & Camera & \multirow{2}{*}{R50} & 850$\times$450 & 59.3  & 64.9 & 27.0 & 83.0 & 82.8 & 38.3 & 27.4 & 44.9 & 24.0 & 55.4 & 73.6 & 91.7 & 60.7 & 59.8 & 61.1 & 78.2 & 76.5  \\ %

    \textbf{OccFormer} (ours) & Camera & & 704$\times$256 & 68.1 & 69.2 & 36.9 & 91.2 & 84.4 & 47.3 & 59.1 & 61.9 & 42.1 & 58.8 & 82.8 & 93.0 & 67.5 & 67.4 & 68.5 & 81.0 & 78.5 \\
    \midrule
    
    BEVFormer~\cite{BEVFormer} & Camera & \multirow{3}{*}{\makecell{R101}} & \multirow{3}{*}{1600$\times$900} & 56.2 & 54.0 & 22.8 & 76.7 & 74.0 & 45.8 & 53.1 & 44.5 & 24.7 & 54.7 & 65.5 & 88.5 & 58.1 & 50.5 & 52.8 & 71.0 & 63.0  \\
    
    TPVFormer~\cite{tpvformer} & Camera & & & 68.9  & 70.0 & 40.9 & 93.7 & 85.6 & 49.8 & 68.4 & 59.7 & 38.2 & 65.3 & 83.0 & 93.3 & 64.4 & 64.3 & 64.5 & 81.6 & 79.3  \\ %

    \textbf{OccFormer} (ours) & Camera & & & 70.4 & 70.3 & 43.8 & 93.2 & 85.2 & 52.0 & 59.1 & 67.6 & 45.4 & 64.4 & 84.5 & 93.8 & 68.2 & 67.8 & 68.3 & 82.1 & 80.4\\
    
    \bottomrule
    \end{tabular}
    \label{tab:nusc_lidarseg_val}
\end{table*}
\begin{table*}
    \footnotesize
    \setlength{\tabcolsep}{0.0035\linewidth}
    \caption{\textbf{Detailed Comparison between sampling methods on SemanticKITTI~\cite{behley2019semantickitti} validation set.}}
    \newcommand{\classfreq}[1]{{~\tiny(\semkitfreq{#1}\%)}}  %
    \centering
    \begin{tabular}{l|c|c c c c c c c c c c c c c c c c c c c|c}
        \toprule
        & SC & \multicolumn{20}{c}{SSC} \\
        Sampling Method & IoU
        & \rotatebox{90}{\textcolor{road}{$\blacksquare$} road\classfreq{road}} 
        & \rotatebox{90}{\textcolor{sidewalk}{$\blacksquare$} sidewalk\classfreq{sidewalk}}
        & \rotatebox{90}{\textcolor{parking}{$\blacksquare$} parking\classfreq{parking}} 
        & \rotatebox{90}{\textcolor{other-ground}{$\blacksquare$} other-ground\classfreq{otherground}} 
        & \rotatebox{90}{\textcolor{building}{$\blacksquare$} building\classfreq{building}} 
        & \rotatebox{90}{\textcolor{car}{$\blacksquare$} car\classfreq{car}} 
        & \rotatebox{90}{\textcolor{truck}{$\blacksquare$} truck\classfreq{truck}} 
        & \rotatebox{90}{\textcolor{bicycle}{$\blacksquare$} bicycle\classfreq{bicycle}} 
        & \rotatebox{90}{\textcolor{motorcycle}{$\blacksquare$} motorcycle\classfreq{motorcycle}} 
        & \rotatebox{90}{\textcolor{other-vehicle}{$\blacksquare$} other-vehicle\classfreq{othervehicle}} 
        & \rotatebox{90}{\textcolor{vegetation}{$\blacksquare$} vegetation\classfreq{vegetation}} 
        & \rotatebox{90}{\textcolor{trunk}{$\blacksquare$} trunk\classfreq{trunk}} 
        & \rotatebox{90}{\textcolor{terrain}{$\blacksquare$} terrain\classfreq{terrain}} 
        & \rotatebox{90}{\textcolor{person}{$\blacksquare$} person\classfreq{person}} 
        & \rotatebox{90}{\textcolor{bicyclist}{$\blacksquare$} bicyclist\classfreq{bicyclist}} 
        & \rotatebox{90}{\textcolor{motorcyclist}{$\blacksquare$} motorcyclist\classfreq{motorcyclist}} 
        & \rotatebox{90}{\textcolor{fence}{$\blacksquare$} fence\classfreq{fence}} 
        & \rotatebox{90}{\textcolor{pole}{$\blacksquare$} pole\classfreq{pole}} 
        & \rotatebox{90}{\textcolor{traffic-sign}{$\blacksquare$} traffic-sign\classfreq{trafficsign}} 
        & mIoU\\
        \midrule
        Uniform & 35.41 & \textbf{59.39} & \textbf{30.01} & \textbf{21.16} & 0.18 & \textbf{14.96} & \textbf{25.80} & 7.10 & 0.16 & \textbf{2.69} & 7.94 & 18.77 & 2.43 & 30.14 & 0.00 & 0.00 & 0.00 & \textbf{6.29} & 3.53 & 0.00 & 12.13 \\
        Class-Guided & \textbf{36.50} & 58.85 & 26.88 & 19.61 & \textbf{0.31} & 14.40 & 25.09 & \textbf{25.53} & \textbf{0.81} & 1.19 & \textbf{8.52} & \textbf{19.63} & \textbf{3.93} & \textbf{32.62} & \textbf{2.78} & \textbf{2.82} & 0.00 & 5.61 & \textbf{4.26} & \textbf{2.86} & \textbf{13.46} \\
        \bottomrule
    \end{tabular}\\
    \label{table:compare_sampling_method}
\end{table*}

\section{More Experiments}
\subsection{LiDAR Segmentation Results}
\label{sec:appendix_nusc_lidarseg_val}

In~\cref{tab:nusc_lidarseg_val}, we report the LiDAR segmentation performance on nuScenes validation set with different backbones and input sizes. For the implementation of BEVFormer for LiDAR segmentation, we follow the settings from TPVFormer~\cite{tpvformer}. When ResNet-50~\cite{resnet} is taken as the backbone network, OccFormer with smaller input sizes can notably outperform TPVFormer. When the larger backbone and input sizes are adopted, the advantage of OccFormer is reduced possibly due to the saturation of vision-based methods. Besides, OccFormer is the first method to achieve 70\%+ mIoU for LiDAR segmentation with only multi-view images as input. 

Also, we note that TPVFormer, specifically trained for 3D semantic occupancy, has unsatisfactory performance in LiDAR segmentation. It indicates that the predicted semantic occupancy from TPVFormer, despite reasonable visualizations, fails to contain accurate 3D positions. By contrast, our method can mitigate the problem by jointly solving both predictions.

\subsection{More Ablation Studies} 
\label{sec:appendix_more_ablations}

\paragraph{Detailed Network Structures.}
As shown in~\cref{tab:ablation_encoder_modules}, more detailed structures in the dual-path transformer encoder are ablated. First, the soft weight for fusing the dual-path outputs is removed and we observe an obvious drop in SSC mIoU from 13.46 to 12.73. Second, we remove the windowed attention in the global path, whose weights are shared with the local path, and observe a degradation of around 0.5 mIoU. Finally, we demonstrate the effectiveness of the bottleneck ASPP from the global path, which can extract long-range information for scene-level semantic layouts.

\vspace{-3mm}
\paragraph{Augmentations}
In~\cref{tab:ablation_augmentations}, we ablate the employed augmentation techniques to train OccFormer. Since the attention mechanism, with strong capacities, is prone to over-fitting, these augmentation techniques are essential for reducing over-fitting and improving performance. Also, we find that the 3D augmentation which jointly transforms the 3D feature and the ground-truth semantic occupancy is more important. When it is disabled, the best performance is achieved at the 9th epoch, despite the total training schedule of 30 epochs. 

\begin{table}
    \centering
    \setlength{\tabcolsep}{0.01\linewidth}
    \caption{Ablation study on encoder modules.}
    \vspace{-2mm}
    \begin{tabular}[b]{c|cc}
        \toprule
        Method & IoU$ \uparrow$ & mIoU$ \uparrow$\\
        \midrule
        OccFormer & \textbf{36.50} & \textbf{13.46} \\
        w.o. soft sum. & 35.83 & 12.73 \\
        w.o. shared attn. & 36.28 & 12.93 \\
        w.o. ASPP & 36.12 & 12.92 \\
        \bottomrule
    \end{tabular}
    \vspace{-2mm}
    \label{tab:ablation_encoder_modules}
\end{table}

\begin{table}
    \centering
    \setlength{\tabcolsep}{0.01\linewidth}
    \caption{Ablation study on augmentations.}
    \vspace{-2mm}
    \begin{tabular}[b]{cc|cc}
        \toprule
        Image Aug. & 3D Aug. & IoU$ \uparrow$ & mIoU$ \uparrow$\\
        \midrule
        \cmark & & 36.37 & 12.72 \\
        & \cmark & 35.73 & 12.94 \\
        \cmark & \cmark & \textbf{36.50} & \textbf{13.46} \\
        \bottomrule
    \end{tabular}
    \vspace{-2mm}
    \label{tab:ablation_augmentations}
\end{table}

\subsection{Analysis}
\label{sec:appendix_analysis}

\begin{figure}[t]
\centering
\includegraphics[width=\linewidth]{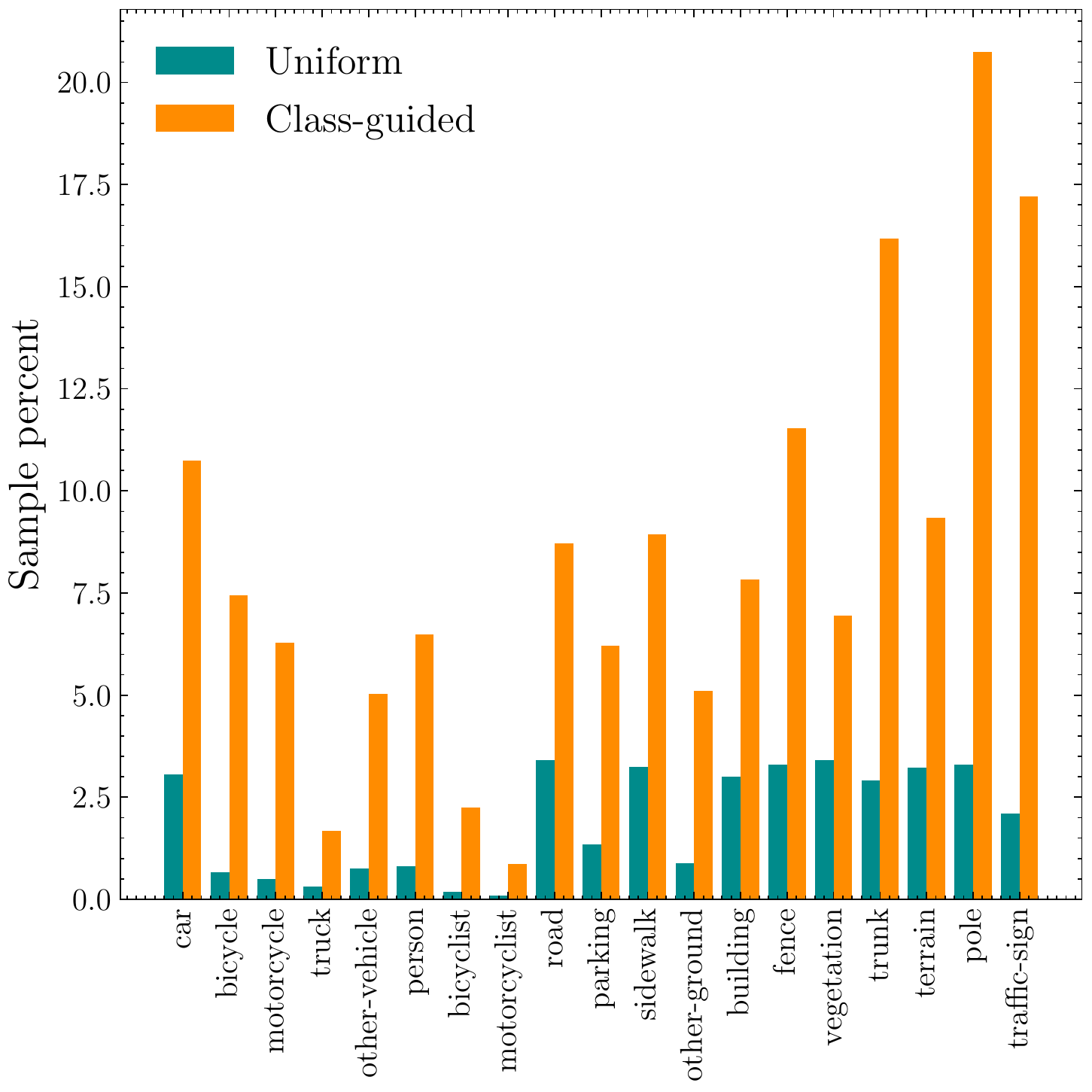}
\vspace{-3mm}
\caption{Comparisons of the uniform sampling and the proposed class-guided sampling. The sample percent is computed as the average sample ratio with 10k times of sampling. The class-guided sampling can significantly improve the quality of supervision.}
\vspace{-4mm}
\label{fig:sample_ratios}
\end{figure}

\paragraph{Class-Guided Sampling.} 
Since the 3D feature volume contains a vast number of positions to supervise, a more effective sampling method is required to enable efficient training. As shown in~\cref{fig:sample_ratios}, the proposed class-guided sampling can greatly improve the supervision signals for rare classes. Quantitatively, the class-wise comparison between uniform sampling and our class-guided sampling is presented in~\cref{table:compare_sampling_method}. Despite minor degradation in larger classes including road, sidewalk, and parking, the class-guided sampling demonstrates a remarkable boost in fewer classes, such as truck, person, bicyclist, and traffic sign. The different patterns from different sampling methods also offer an approach for the model ensemble.

\begin{figure*}[t]
\centering
\includegraphics[width=0.95\linewidth]{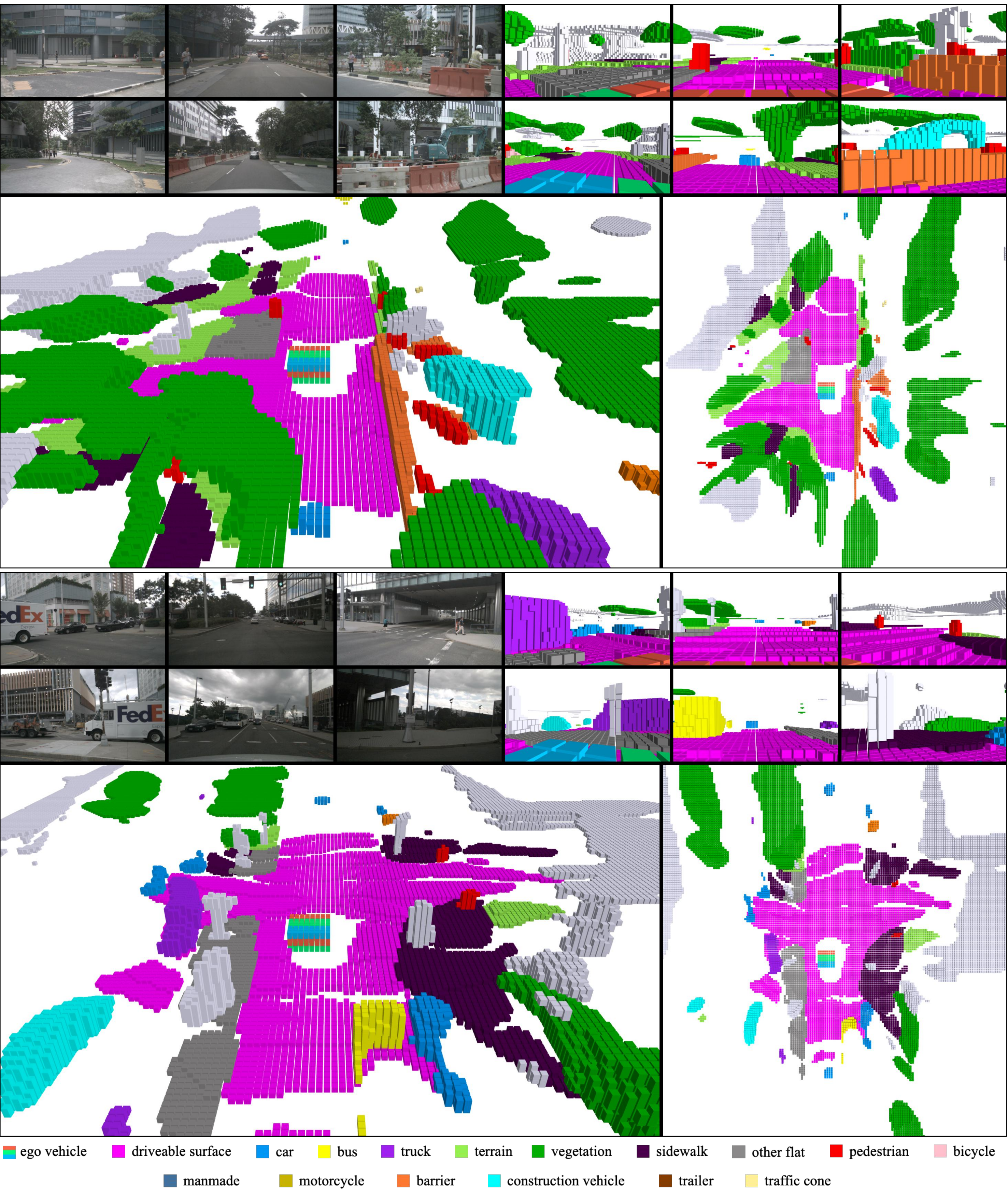}
\vspace{-2mm}
\caption{
\textbf{Qualitative results on nuScenes validation set.} Two representative samples are selected. For each sample, the input multi-view images are shown on the top left. The predicted semantic occupancy is shown from every camera view (top right), the front overlook (bottom left), and the bird-eye-view (bottom right). The red-green-blue box represents the ego vehicle.}
\label{fig:nusc_supple_visualize}
\vspace{-3mm}
\end{figure*}

\section{More Visualizations}
\label{sec:appendix_visualize}

In~\cref{fig:nusc_supple_visualize}, we provide more qualitative results for 3D semantic occupancy prediction on nuScenes validation set. Though OccFormer takes multi-view 2D images as input and is trained with sparse LiDAR points, it can predict dense results for background classes including vegetation, driveable surface, and building. Also, the foreground objects like cars, pedestrians, and trucks can be located accurately. The predicted 3D semantic occupancy can serve as a comprehensive and fine-grained understanding of the surrounding environment. The video demos on SemanticKITTI and nuScenes datasets are also available at the project page\footnote{https://github.com/zhangyp15/OccFormer}. 



{\small
\bibliographystyle{ieee_fullname}
\bibliography{arxiv}
}

\end{document}